\documentclass[journal]{IEEEtran}
\ifCLASSINFOpdf
  \usepackage[pdftex]{graphicx}
  \graphicspath{{../pdf/}{../jpeg/}}
  \DeclareGraphicsExtensions{.pdf,.jpeg,.png}
\else
  \usepackage[dvips]{graphicx}
  \graphicspath{{../eps/}}
  \DeclareGraphicsExtensions{.eps}
\fi
\usepackage{amsmath}
\usepackage{amssymb}
\usepackage{mathrsfs}
\usepackage{cases}
\usepackage{algorithmic}

\usepackage{multirow}

\newtheorem{theorem}{Theorem}
\newtheorem{definition}{Definition}

\begin{document}

\title{Deep-ESN: A Multiple Projection-encoding Hierarchical Reservoir Computing Framework}

\author{Qianli Ma, \emph{Member}, \emph{IEEE},
        Lifeng Shen,
        Garrison W. Cottrell% <-this % stops a space
\thanks{Qianli Ma is with the School of Computer Science and Engineering, South China University of Technology, Guangzhou, 510006, China, %and also with Guangdong Key Laboratory of Big Data Analysis and Processing, Guangzhou, 510006, China 
(e-mail: qianlima@scut.edu.cn).}% <-this % stops a space
\thanks{Lifeng Shen is with the School of Computer Science and Engineering, South China University of Technology, Guangzhou, 510006, China (e-mail: scuterlifeng@foxmail.com)}
\thanks{Garrison W. Cottrell is with the Department of Computer Science and Engineering, University of California, San Diego, CA 92093, USA (e-mail: gary@ucsd.edu).}% <-this % stops a space
}


\markboth{Journal of \LaTeX\ Class Files,~Vol.~14, No.~8, August~2017}%
{Shell \MakeLowercase{\textit{et al.}}: Bare Demo of IEEEtran.cls for IEEE Journals}



\maketitle

\begin{abstract}
As an efficient recurrent neural network (RNN) model, reservoir computing (RC) models, such as Echo State Networks, have attracted widespread attention in the last decade. However, while they have had great success with time series data \cite{Lukosevicius:2009,Jaeger2004Harnessing}, many time series have a multiscale structure, which a  single-hidden-layer RC model may have difficulty capturing. In this paper, we propose a novel hierarchical reservoir computing framework we call Deep Echo State Networks (Deep-ESNs). The most distinctive feature of a Deep-ESN is its ability to deal with time series through hierarchical projections. Specifically, when an input time series is projected into the high-dimensional echo-state space of a reservoir, a subsequent encoding layer (e.g., a PCA, autoencoder, or a random projection) can project the echo-state representations into a lower-dimensional space. These low-dimensional representations can then be processed by another ESN. By using projection layers and encoding layers alternately in the hierarchical framework, a Deep-ESN can not only attenuate the effects of the collinearity problem in ESNs, but also fully take advantage of the temporal kernel property of ESNs to explore multiscale dynamics of time series. To fuse the multiscale representations obtained by each reservoir, we add connections from each encoding layer to the last output layer. Theoretical analyses prove that stability of a Deep-ESN is guaranteed by the echo state property (ESP), and the time complexity is equivalent to a conventional ESN. Experimental results on some artificial and real world time series demonstrate that Deep-ESNs can capture multiscale dynamics, and outperform both standard ESNs and previous hierarchical ESN-based models.

\end{abstract}


% \fi
\IEEEpeerreviewmaketitle

% !TeX encoding = UTF-8
\section{Introduction}

\IEEEPARstart{R}{eservoir} computing (RC) \cite{Lukosevicius:2009} is a framework for creating recurrent neural networks (RNNs) efficiently that have a ``reservoir'' of dynamics that can be easily tapped to process complex temporal data.  By using fixed input and hidden-to-hidden weights, reservoir computing avoids the laborious process of gradient-descent RNN training, yet can achieve excellent performance in nonlinear system identification \cite{soh2015spatio,scardapane2016decentralized}, signal processing \cite{xia2011augmented} and time series prediction \cite{Jaeger2004Harnessing,li2012chaotic,xu2016adaptive,ma2013modular,soriano2015delay}.

The two most well-known RC models are Liquid State Machines and Echo State Networks \cite{Jaeger2004Harnessing,Jaeger2001The,Maass-2002-LSNs}. Liquid State Machines have continuous dynamics and spiking neurons, while Echo State Networks (ESNs) use discrete dynamics and rate-coded neurons. Due to their relative simplicity, ESNs are more widely used. An ESN usually consists of three components: an input layer, a large RNN layer (called the \emph{reservoir}) and a linear output layer. The weights in the input and reservoir layers are randomly initialized and fixed during the learning stage. The reservoir is initialized with sparse connections and a constraint on the spectral radius of the weight matrix that guarantees rich, long-term dynamics \cite{Lukosevicius:2009}. The reservoir of an ESN can be viewed as a nonlinear temporal kernel, which can map sequences of inputs into a high-dimensional space, and learning is reduced to linear regression from the reservoir to the output. Hence, ESNs are a powerful tool for analysis of dynamic data by simplifying the training process.

However, hierarchical multiscale structures naturally exist in temporal data \cite{chung2016hierarchical}, and a single RNN can have difficulty dealing with input signals that require explicit support for multiple time scales \cite{NIPS2013_5166}. Therefore, to quote Herbert Jaeger, ``\textit{A natural strategy to deal with multiscale input is to design hierarchical information
processing systems, where the modules on the various levels in the
hierarchy specialize on features on various scales.}'' \cite{Jaeger2007Discovering}.

In recent years, several hierarchical RNN architectures have been proposed \cite{chung2016hierarchical,NIPS2013_5166,fernandez2007sequence,Deep_Bidirectional,graves2013speech,pascanu2013construct}. However, the lengthy process of training deep RNNs is still a practical issue \cite{pascanu2013construct}. Hence, constructing a hierarchical ESN is an attractive approach to this problem as the training is trivial. Nevertheless, constructing a hierarchical ESN-based model while maintaining the stability and echo state property (ESP) of the hierarchical system is still a challenging issue.

%learning multi-scales features is an important topic in the filed of deep learning. For example, deep autoencoder networks can capture multi-scales low-dimensional codes (?) in a unsupervised way \cite{Hinton2006Reducing}, and convolutional neural networks (CNNs) extract high abstract representations by utilizing multiple convolution and pooling layers \cite{Krizhevsky:2012CNN}, and
%deep recurrent neural networks, and deep bidirectional LSTM, etc. They both emphasize encoding features along their layers.
%Additionally, for traditional recurrent neural networks like LSTM, there also have corresponding hierarchical variants, e.g.,

The first attempt to develop a hierarchical ESN is the dynamical feature discoverer (DFD) model proposed by Jaeger~\cite{Jaeger2007Discovering}. The main idea is that by stacking multiple reservoirs, the outputs of a higher level in the hierarchy serve as coefficients of mixing (or voting on) outputs from a lower one \cite{Lukosevicius:2009}. It learns the outputs from each reservoir simultaneously by a gradient-based algorithm, which increases the computational complexity compared to the linear regression training of the original ESN. Subsequently, Triefenbach et al. \cite{triefenbach2010phoneme,triefenbach2013acoustic,cascaded2014} explored cascaded ESNs to obtain multi-level phonetic states, and used Hidden Markov Models (HMMs) as the phoneme language model for speech recognition. The cascaded ESN feeds outputs of the previous reservoir into the next one in a supervised way and trains output weights layer by layer.

In order to study the memory properties of deep ESNs, Gallicchio and Micheli \cite{gallicchio2016deep} performed an empirical analysis of deep ESNs with leaky integrator units. They constructed an artificial random time series out of 10 1-out-of-N inputs, and constructed a second time series by adding a ``typo'' partway through the series. The goal was to measure how long the representation differed between the original series and the one with a typo. They found that varying the integration parameter, slowing the integration through the stacks lead to very long memory. It is unclear how well this will generalize to realistic time series, but it is an interesting observation.

Most recently, the multilayered echo-state machine (MESM) \cite{MESM} was proposed to pipeline multiple same-size reservoirs. Each reservoir uses the same internal weight matrix, and are connected by subsequent layers by the same coupling weight matrix. As with standard ESNs, the output weights of the last reservoir are the only trainable parameters. They proved that this architecture satisfies the Echo-State Property, and did a rigorous comparison of its performance compared to standard ESNs on multiple datasets. They achieved better results compared to standard ESNs. However, by setting the reservoir size in each layer to the same size as \cite{MESM,gallicchio2016deep} did, the hierarchical models do not take advantage of the high-dimensional projection capacity since the representations generated by the first reservoir will be projected into a state space with the same dimension of the first one.

Apart from above-mentioned ESN-based hierarchy, some work tries to take advantage of a random-static-projection technique, the Extreme Learning Machine (ELM) \cite{huang2015trends}, to augment the nonlinearity in reservoirs. One such model is the $\varphi$-ESN \cite{gallicchio2011architectural}, which is a two-layer model adding a static feed-forward layer of ELM to a reservoir. The main idea is to use the ELM to increase the dimensionality over that of the reservoir, in order to increase separability. The other is R$^2$SP \cite{Butcher201376,butcher2010extending} which is very similar to the $\varphi$-ESN, adding two ELMs to encode inputs and reservoir states respectively. Their results showed that employing some static feed-forward networks as intermediate layers will obtain more nonlinear computational power in the reservoir.

%RC provides a quick and efficient training for RNNs when compared to methods based on gradient descent.
% A similar work is the deep-stacking ESN \cite{Deng2014}, which constructs reservoirs in a pipelined way as cascaded ESN. However, it adopts back-propagation through time (BPTT)[cited?] to optimize all parameters in networks which is not as efficient as ESN’s regression way.

%And the votes is formulated by features and votes of the upper layers. Other layers run in the same manner. Finally, all loss information is propagated in a bottom-to-top way. However, this model can't be argued a hierarchical reservoir-computing in real sense.

As mentioned before, directly stacking several reservoirs is not sufficient to built a efficient hierarchical ESN-based model. On one hand, it is well known that the most important property of ESN is the high-dimensional projection capacity of reservoir. If connecting several same size reservoirs together as \cite{gallicchio2016deep} and \cite{MESM} did, usually only a small advantage is obtained over one layer, and the MESM model did not appear to gain anything beyond two layers. On the other hand, due to the reservoir usually being a big randomly sparsely connected RNN, when the input data are expressed as echo-state representations by the reservoir, some of the representations tend to be redundant. This is called the collinearity problem of ESNs \cite{li2012chaotic,xu2016adaptive}. Therefore, in a hierarchical ESN, encoding the high-dimensional echo-state representations of the previous reservoir to proper low-dimensional data is vital for playing a role of projection in the next reservoir. Furthermore, although \cite{gallicchio2016deep} explored the time-scale hierarchical differentiation among layers by providing input data to every intermediate layer, most existing hierarchical ESN-based models fail to fuse multiple time-scale features at the last layer.

%there still are no any efficient works to develop a hierarchical reservoir-computing frameworks to learn the multi-scales characteristics of time series data. And to the best of our knowledge, there also are no any successful attempts to bridge the shallow paradigm reservoir computing and the main-stream deep learning.

%MESM isn't still ideal hierarchical reservoir-computing framework. The main reasons are:
%\begin{itemize}
%	\item
%%It excutes multiple state transitions in time direction and along various layers, while it is unclear why there many transitions along the layers;
%	\item
%It can't use multi-scales features as there are no any learned encoders in each layer;
%	\item It does not fully utilizes the strength of high-dimensional projection in reservoir computing.
%\end{itemize}
%Since ELM also emphasize high-dimensional random projection and training the only output weights, it can be viewed as a non-recurrent RC network.

To address these problems, this paper proposes a novel multiscale deep reservoir computing framework we call Deep Echo State Networks (Deep-ESN). A distinctive feature of the Deep-ESN is that it uses a projection layer and an encoding layer alternately in the hierarchical framework. More specifically, when an input time series is projected into the echo-state space of a reservoir, a subsequent encoding layer (e.g., PCA, autoencoder, or random projection) receives the echo states of the previous reservoir as input and encodes the high-dimensional echo-state representations into a lower dimension. Then these low-dimensional representations are once again projected into the high-dimensional space of the following reservoir. By this multiple projection-encoding method, the Deep-ESN can not only attenuate the effects of the collinearity problem in ESNs, but also can fully take advantage of the temporal kernel property of each reservoir to represent the multiscale dynamics of the time series. Moreover, to integrate the multiscale representations, we add connections from each encoding layer to the last output layer. The stability of the Deep-ESN is guaranteed by the ESP, and the run-time complexity is equivalent to a standard ESN. The main contributions of this paper can be summarized as follows.
\begin{enumerate}
	\item We propose a novel multiple projection-encoding deep reservoir computing framework called Deep-ESN, which bridges the gap between reservoir computing and deep learning.
	\item By unsupervised encoding of echo states and adding direct information flow from each encoder layer to the last output layer, the proposed Deep-ESN can not only obtain multiscale dynamics, but also dramatically attenuate the effects of collinearity problem in ESNs.
	\item In a theoretical analysis, we analyze the stability and computational complexity of Deep-ESN. We also verify that the collinearity problem is alleviated with small condition numbers, and that each layer of the Deep-ESN can capture various dynamics of time series.
%applying recurrence quantification analysis (RQA) \cite{marwan2007recurrence,eroglu2014entropy,bianchi2016investigating}
	\item Compared with the several RC hierarchies, Deep-ESN achieves better performances on well-known chaotic dynamical system tasks and some real world time series.
%(e.g., $\varphi$-ESN, R$^2$SP and MESM)	
\end{enumerate}

The rest of this paper is organized as follows. In Section II, we introduce typical ESN architecture and its properties and related hyper-parameters. In Section III, we describe the details of the proposed Deep-ESN and then we conduct stability and computational complexity analysis. After that, we report our experimental results in Section IV. We also give the structure analysis, collinearity analysis, and dynamic analysis of Deep-ESN. Finally, we give the discussion and conclusions in Section V.
% These encoders all are trained in unsupervised way, which simplifies the construction of whole framework.

%in which we cascades multiple reservoirs with specific encoders in pipeline way. After them, a basic hierarchical reservoir network has been built.

\section{Echo State Network}

\begin{figure}[!t]
	\centering
	\includegraphics[width=0.25\textwidth]{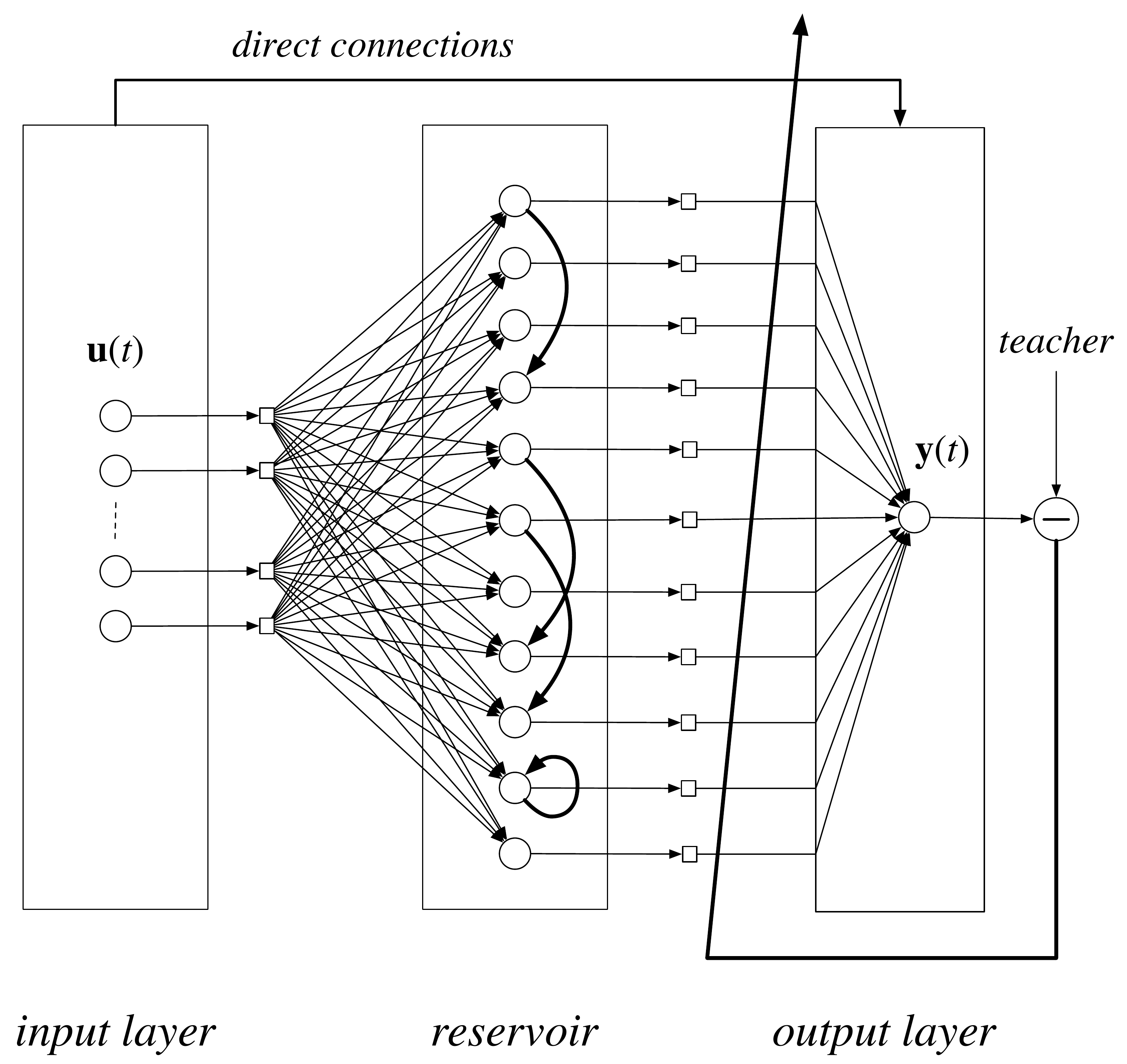}
	\caption{Architecture of the basic ESN, which consists of three components: input layers with $D$ neurons, large resrevoir with $N$ neurons and the last output with $L$ neurons ( where $L=1$). The blueberry links denote the fixed sparse self-connections in reservoir.}
	\label{esn}
\end{figure}

\subsection{ESN Architecture}

Echo state network is a recurrent neural network consisting of three basic components: an input layer, a large recurrent layer (called the \emph{reservoir}) with fixed sparse hidden-to-hidden connections and an output layer. The general architecture of an ESN is illustrated in Fig.\ref{esn}.

Let $D$, $N$ and $L$ denote the numbers of neurons in the input layer, the reservoir and the output layer. The weights of input-to-reservoir and reservoir-to-reservoir are collected by an $N$-by-$D$ matrix $\mathbf{W}^{in}$ and an $N$-by-$N$ matrix $\mathbf{W}^{res}$. The weights from input-to-output and reservoir-to-output are cascaded into a single $L$-by-$(D+N)$ matrix $\mathbf{W}^{out}$. Among these, $\mathbf{W}^{in}$ is randomly initialized from a uniform distribution $[-1,1]$, and $\mathbf{W}^{res}$ is defined in (\ref{eq:W_res}). They are fixed during the training stage, and only $\mathbf{W}^{out}$ needs to be adapted.

An ESN is trained by supervised learning. Two steps are involved. The first one is to map $D$-dimensional inputs $\mathbf{u}$ into a high-dimensional reservoir state space, driving the reservoir to obtain the echo states $\mathbf{x}$. The other step is to learn the output matrix $\mathbf{W}^{out}$ by simple regression techniques. Here we introduce an ESN with leaky-integrator neurons proposed by Jaeger \cite{jaeger2007optimization}, which is also adopted by \cite{MESM,gallicchio2016deep}. The mathematical formulations for the entire system are as follows:
\begin{equation}
	\mathbf{z}(t)= f(\mathbf{W}^{res}\mathbf{x}(t)+\mathbf{W}^{in}\mathbf{u}(t+1))
	\label{eq:project}
\end{equation}
\begin{equation}
	\mathbf{x}(t+1)=(1-\gamma)\mathbf{x}(t)+\gamma\mathbf{z}(t)
\label{eq:update}
\end{equation}
\begin{equation}
	\mathbf{y}(t+1)=f^{out}(\mathbf{W}^{out}[\mathbf{x}(t+1); \mathbf{u}(t+1)])
\label{eq:train}	
\end{equation}
where $\mathbf{u}$, $\mathbf{x}$ and $\mathbf{y}$ denote the inputs, the reservoir states and the outputs, respectively.  $f(\cdot)$ is the non-linear activation function in reservoir (usually $tanh(\cdot)$) and $f^{out}$ is the activation function in output (usually $identity(\cdot)$). $t$ denotes the time step. $\gamma$ in (\ref{eq:update}) denotes the leak rate which is used for integrating the states of the previous time step with the current time step.

There are three main characteristics that distinguish an ESN from other RNNs:
\begin{enumerate}
	\item ESN adopts a random high-dimensional projection method to capture the dynamics of the inputs, which has a similar function to that of the kernel in kernel-based learning methods \cite{Lukosevicius:2009};
	\item The reservoir is the core of whole system, which consists of large number (typically 100-1000D) of sparsely connected neurons, and none of the weights in the reservoir are trained.
	\item The output signals are the linear combinations of the echo states of the reservoir, and simple linear regression algorithms can compute the linear readout layer weights.
\end{enumerate}

Therefore, compared with other RNNs, training an ESN is both simple and fast. Moreover, it does not get stuck in local minima, which endows it with high computational capabilities for modeling underlying dynamical system of time series.

\subsection{Hyperparameters and Initializations}

The important hyperparameters used for initializing an ESN are \textit{IS} - the input scaling, \textit{SR} - the spectral radius, $\alpha$ - the sparsity and the aforementioned leak rate $\gamma$.
\begin{enumerate}
\item \textit{IS} is used for the initialization of the matrix \(\textbf{W}^{in}\): the elements of \(\textbf{W}^{in}\) obey the uniform distribution of -\(IS\) to \(IS\). %We set \(IS\) to 0.1.
\item \textit{SR} is the spectral radius of \(\textbf{W}^{res}\), given by
\begin{equation}
\mathbf{W}^{res}=SR\cdot\frac{\mathbf{W}}{\lambda_{max}(\mathbf{W})}
\label{eq:W_res}	
\end{equation}
where $\lambda_{max}(\mathbf{W})$ is the largest eigenvalue of matrix $\mathbf{W}$ and the elements of $\mathbf{W}$ are generated randomly from $[-0.5, 0.5]$. To satisfy the Echo State Property (ESP) \cite{Jaeger2001The,Yildiz2012ESP}, \textit{SR} should be set smaller than 1. This is a necessary condition of ESN stability. The ESP will be discussed in more detail later.
\item $\alpha$ denotes the proportion of non-zero elements in $\mathbf{W}^{res}$. We set $\alpha$ to 0.1.
\end{enumerate}

In short, ESNs have a very simple training procedure, and due to the high-dimensional projection and highly sparse connectivity of neurons in the reservoir, it has abundant non-linear echo states and short-term memory, which are very useful for modeling dynamical systems. However, a single ESN can not deal with input signals that require complex hierarchical processing and it cannot explicitly support multiple time scales. In the next section, we will propose a novel deep reservoir computing framework to resolve this issue.

\section{Deep Echo State Network}
In this section, the details of the proposed Deep-ESN will be described, as well as the related analysis of stability and computational complexity.

\begin{figure*}[!t]
	\centering
	\includegraphics[width=0.75\textwidth]{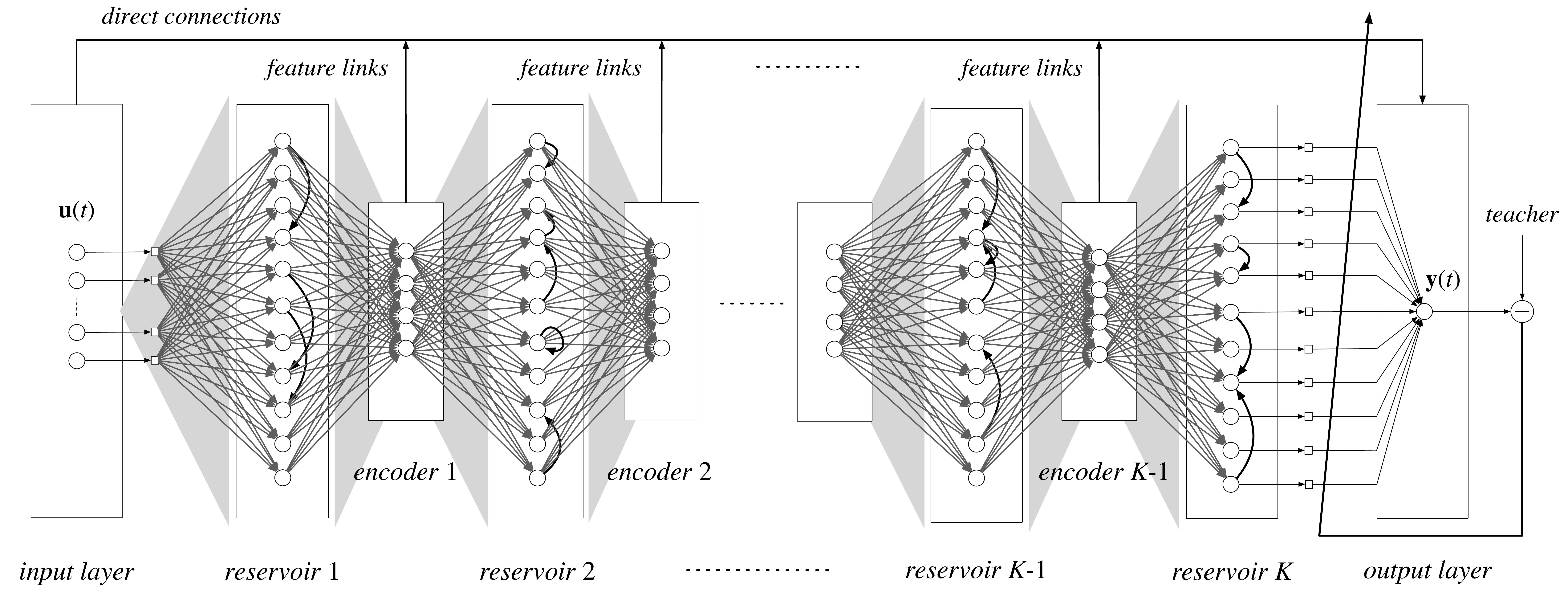}
	\caption{Architecture of the proposed Deep-ESN with $K$ reservoirs and $(K-1)$ encoders. The direct connections from inputs to output layer and the feature links from each encoder to output layer will be cascaded and connected into the output neurons.}
	\label{drn}
\end{figure*}

Although the main idea of hierarchical ESN-based models is to capture multiscale dynamics of time series by constructing a deep architecture, there are three main features of our Deep-ESN that contrast with previous approaches:
%, which is consistent with the goal of Jaeger's works \cite{Jaeger2007Discovering}
\begin{itemize}
	\item \textit{Multiple Projection-Encoding}: Instead of directly stacking multiple ESNs, Deep-ESN uses the encoder layer between reservoirs. In this way, the DeepESN can not only obtain abundant multiscale dynamical representations of inputs by fully taking advantage of high-dimensional projections, but also solves the collinearity problem in ESNs.
	\item \textit{Multiscale feature fusion}: In order to better fuse multiscale dynamical representations captured by each reservoir, we add connections (called feature links) from each encoding layer to the last output layer.
%Consider reservoir is a wide structure and avoid later reservoir grows too huge, it is reasonable to imply feature encoders to compress size of the former reservoir.	
	\item \textit{Simplicity of Training}: Unlike some previous hierarchical ESN-based models, training the whole model layer by layer, the only trainable layer of the Deep-ESN is the last output layer, which retains the efficient computation of RC without relying on gradient-propagation algorithms.
\end{itemize}

\subsection{Deep-ESN Framework}
The architecture of the Deep-ESN is illustrated in Fig.\ref{drn}. Its hidden layers consist of $K$ reservoirs and $(K-1)$ encoders. To avoid confusion, we define this Deep-ESN as a $K$-layer network. Let $N^{(i)}$ and $M^{(j)}$ denote the number of neurons in the $i$-th reservoir layer and the $j$-th encoder layer, where $i=1,\dots,K$ and $j=1,\dots,K-1$. The $T$-length training series inputs are denoted as  $\mathbf{u}=[\mathbf{u}(1),\mathbf{u}(2),\dots,\mathbf{u}(T)]$ and teacher signals are denoted as $\mathbf{d}=[\mathbf{d}(1),\mathbf{d}(2),\dots,\mathbf{d}(T)]$,  where $\mathbf{u}(t)\in\mathbb{R}^D$, $\mathbf{d}(t)\in\mathbb{R}^L$ at each time step $t$. Activations (updated states) in $i$-th layer reservoir and $j$-th layer encoder are denoted as $\mathbf{x}_{res}^{(i)}(t)$ and $\mathbf{x}_{enc}^{(j)}(t)$ respectively, where $i=1,\dots,K$, $j=1,\dots,K-1$ and $t= 1,\dots,T$. Further, We use $\mathbf{W}^{in(i)}$ to collect the input weights of $i$-th reservoir, $\mathbf{W}^{res(i)}$ to collect the recurrent weights and $\mathbf{W}^{enc(j)}$ to collect the input weights of $j$-th encoder. Matrix $\mathbf{W}^{out}$ has all the weights of the direct connections, the output weights from the last reservoir and the weights of feature links. In this way, the formulation details of our proposed Deep-ESN can be presented as follows.

For the $i$-th reservoir ($i=1,\dots,K$), its high-dimensional states can be obtained by
\begin{equation}
\mathbf{z}^{(i)}(t)=
f(\mathbf{W}^{res(i)}\mathbf{x}_{res}^{(i)}(t)+\mathbf{W}^{in(i)}\mathbf{x}_{in}^{(i)}(t+1))
\label{eq:project_drn}
\end{equation}
\begin{equation}
\mathbf{x}_{res}^{(i)}(t+1)=(1-\gamma)\mathbf{x}_{res}^{(i)}(t)+\gamma\mathbf{z}^{(i)}(t)
\label{eq:update_drn}
\end{equation}
where $\mathbf{x}_{in}^{(i)}(t+1)$ denotes the inputs of $i$-th reservoir. When $i$ equals one, we have $\mathbf{x}_{in}^{(1)}(t+1)=\mathbf{u}(t+1)$. When $i$ is greater than one, we have $\mathbf{x}_{in}^{(i)}(t+1)=\mathbf{x}_{enc}^{(i-1)}(t)$, which means the inputs of $i$-th reservoir are the output of the $(i-1)$-th encoder.
%{\Large GWC: $x_{in}$ can't be $x_{res}$ - it has to be the output of the encoder, i.e., $x_{enc}$ You need to have equation 9 earlier.}
For simplicity, we use an operator $\mathcal{F}_i$ to denote the high-dimensional projection (\ref{eq:project_drn}) and the update step (\ref{eq:update_drn}). That is
\begin{equation}
\mathbf{x}_{res}^{(i)}(t+1) =
\mathcal{F}_i(\mathbf{x}_{res}^{(i)}(t), \mathbf{x}_{in}^{(i)}(t+1))
\end{equation}

Given the states of the previous reservoir, we can use an unsupervised dimension reduction technique $\mathcal{T}$ to encode them and produce encoded features. Thus, the encoding procedure of the $j$-th encoder ($j=1,\dots,K-1$) can be formulated as
\begin{equation}
\mathbf{x}_{enc}^{(j)}(t) = \mathcal{T}(\mathbf{x}_{res}^{(j)}(t))
\label{eq:encoder_tao}
\end{equation}
Further, we can instantiate $\mathcal{T}$ in (\ref{eq:encoder_tao}) as
\begin{equation}
\mathcal{T}(\mathbf{x}_{res}^{(j)}(t)) = f_{enc}(\mathbf{W}^{enc(j)}\mathbf{x}_{res}^{(j)}(t))
\label{eq:encoder_instance}
\end{equation}
where $f_{enc}(\cdot)$ is the activation function of the encoder. When $f_{enc}(\cdot)$ is the identity function, $\mathcal{T}$ is a linear dimensionality reduction technique. The choices of $\mathcal{T}$ will be introduced later.

According to above description, we can obtain the state representations of the last reservoir by
\begin{equation}
\mathbf{x}_{res}^{(K)}(t+1) = \mathcal{F}_{K} \circ\mathcal{H}_{K-1}\circ\dots\circ\mathcal{H}_{1}(\mathbf{u}(t+1))
\end{equation}
where $\mathcal{H}_{j}=\mathcal{T}_{j}\circ\mathcal{F}_{j}$ and the symbol $\circ$ denotes a composition operator.

Unlike the traditional ESNs, the Deep-ESN incorporates additional middle-layer encoded features into the last output layer. Outputs of the Deep-ESN at time $t+1$ can be computed by
\begin{equation}
\mathbf{y}(t+1)=f^{out}(\mathbf{W}^{out}\mathbf{M}(t+1))
\label{eq:drn_output}
\end{equation}
where $\mathbf{M}(t+1)$ is:
\begin{equation}
\mathbf{M}(t+1)=[\underbrace{\mathbf{x}_{res}^{(K)}(t+1)^T}_{\text{A}}, \underbrace{\mathbf{u}(t+1)^T}_{\text{B}}, \underbrace{\{\mathbf{x}_{enc}^{(1,\dots,K-1)}(t+1)^T\}}_{\text{C}}]^T
\label{eq:collect}
\end{equation}
where A is the echo states of the last reservoir, B is the inputs along with direct connections, and C is the multiscale representations along with the feature links. Rewriting (\ref{eq:drn_output}) in matrix form, we have:
\begin{equation}
\mathbf{Y}=f^{out}(\mathbf{W}^{out}\mathbf{M})
\label{eq:drn_output_matrix}
\end{equation}
%{\Large The equation above is missing $f^{out}$}
where $f^{out}$ is element-wised output activation function, $\mathbf{Y}=[\mathbf{y}(1),\mathbf{y}(2),\dots,\mathbf{y}(T)]$ and $\mathbf{M}=[\mathbf{M}(1),\mathbf{M}(2),\dots,\mathbf{M}(T)]$.

Additionally, if the teacher signals matrix is defined as $\mathbf{T}=[\mathbf{d}(1),\mathbf{d}(2),\dots,\mathbf{d}(T)]$, we have the squared error loss of the whole system:
\begin{equation}
E(\mathbf{W}^{out})\propto\|\mathbf{Y}-\mathbf{T}\|_2^2
\label{eq:drn_lossfunc}
\end{equation}
which is still a simple regression problem on the parameters $\mathbf{W}^{out}$. Since time series present a high-dimensional form ($T$ is too large), this problem always is over-determined and we adopt ridge-regression with Tikhonov regularization \cite{tikhonov1977solutions} to solve it.
\begin{equation}
	\hat{\mathbf{W}}^{out}=\mathbf{T}\mathbf{M}^T(\mathbf{M}\mathbf{M}^T+\beta\mathbf{I})^{-1}
\label{eq:train_weights}
\end{equation}
where $\beta$ is a small regularization term (here fixed to $10^{-5}$).

\subsection{Choices of Encoders}
To retain the computational advantages of RC, the encoder $\mathcal{T}$ should have low learning cost. Three dimensionality reduction (DR) techniques are used.

1) \textit{Principal Component Analysis} (PCA) is a popular DR statistical method. PCA adopts an orthogonal base transformation to project the observations into a linearly uncorrelated low-dimensional representation where the selected orthogonal bases are called \textit{principal components}. In mathematical terms, PCA attempts to find a linear mapping $\mathbf{W}\in \mathbb{R}^{D\times M}$ ($M<D$) that maximizes the following optimization problem:
\begin{eqnarray}
&\mathbf{W}^{\star}=\mathop{\arg\max}_\mathbf{W}{\|\mathbf{W}^T\mathbf{S}_x\mathbf{\mathbf{W}}\|_2}\\
&subject\ to\ \ \mathbf{W}^T\mathbf{W}=\mathbf{I}_M
\end{eqnarray}
where $\mathbf{S}_x=\mathbf{X}\mathbf{X}^T$ is the covariance matrix of zero-mean inputs $\mathbf{X}\in \mathbb{R}^{D\times N}$. $\mathbf{I}_M$ is the $M\times M$ identity matrix. $D$ and $M$ are the original and reduced dimension of $\mathbf{X}$ respectively. The optimal $\mathbf{W}^{\star}$ can be provided by eigenvectors corresponding to the $m$ largest eigenvalues of the covariance matrix $\mathbf{S}_x$. While standard PCA is dominated by the eigenvalue decomposition, and so is $\mathcal{O}(D^2N + D^3)$ \cite{Maaten2007Dimensionality}, there are fast, iterative methods that are $\mathcal{O}(D^2N + D^2ML)$, where $N$ is the number of data points, $M$ is the number of leading eigenvectors required, and $L$ is the number of iterations to converge, which is usually quite small (usually 2-5), and therefore the complexity can be rewritten to $\mathcal{O}(D^2N + D^2M)$~\cite{sharma-fast-pca}.

2) \textit{ELM-based Auto-encoder} (ELM-AE) \cite{6733226,multi_ELM2016} is a recent DR tool based on Extreme Learning Machine (ELM), which is used for simplifying the training of traditional auto-encoders. The main idea is to obtain the hidden random features $\mathbf{H}\in\mathbb{R}^{M\times N}$ by using random weights $\mathbf{W}^0\in\mathbb{R}^{M\times D}$ and bias $\mathbf{b}^0\in\mathbb{R}^{M\times D}$, formulated by
\begin{align}
\begin{gathered}
\textbf{H}=g(\textbf{W}^0\textbf{X}+\textbf{b}^0)
\end{gathered}
\end{align}
where $\mathbf{X}\in \mathbb{R}^{D\times N}$ is the inputs and $g$ denotes the activation function. Then the dimension reduction mappings $\mathbf{W}^{\star}\in\mathbb{R}^{D\!\times\!M}$ can be solved by optimizing the following problems:
\begin{equation}
\mathbf{W}^{\star}=\mathop{\arg\max}_\mathbf{W}{\|\mathbf{W}\mathbf{H}-\mathbf{X}\|_2+\lambda\|\mathbf{W}\|_2}
\label{eq:multiELM}
\end{equation}
where $\lambda$ is the regularization coefficient. This problem is a simple regression problem and can be solved by the pseudo-inverse technique, same as (\ref{eq:train_weights}). Finally, the reduced data $\mathbf{H}_{enc}$ can be represented by $\mathbf{H}_{enc}=(\mathbf{W}^{\star})^T\mathbf{X}$. The computational complexity of (\ref{eq:multiELM}) is $\mathcal{O}(DMN)$.

3) \textit{Random Projection} (RP) is a kind of Monte Carlo method which constructs Lipschitz mappings  \cite{Johnson1984Extensions} to preserve the local separation of high-dimensional data with a high probability \cite{Bingham:2001}. Different RP methods can be realized by various random matrices \cite{Wang2011}. In Deep-ESN, we select the random matrix $\mathbf{W}\in\mathbb{R}^{D\times M}$ designed by Achlioptas \cite{Achlioptas:2001}. The elements $w_{ij}$ of $\mathbf{W}$ are given by this distribution:
\begin{numcases}{w_{ij}=\sqrt{3}\times}
+1,\ with\ probability\ 1/6, \label{w:1}\notag\\
0,\ \ \ with\ probability\ 2/3, \label{w:2}\\
-1,\ with\ probability\ 1/6.\label{w:3}\notag
\end{numcases}
Compared with PCA and ELM-AE, RP has much lower computational expense with $\mathcal{O}(DM)$, where $D$ and $M$ are the original and reduced dimension of $\mathbf{X}$, respectively.

\subsection{Heuristic Optimization of Hyperparameters}
Reasonable setting of the hyperparameters is vital to building a high performance RC network. There are three commonly used strategies: direct method (based on user's experience), grid search and heuristic optimization. The former two strategies are used for general single-reservoir ESNs, but they are unsuitable for Deep-ESNs due to its larger parameter space. Thus, we adopt heuristic optimization to set hyperparameters. The genetic algorithm (GA) is a commonly-used heuristic technique to generate high-quality solutions to optimization and search problems \cite{Mitchell:1996}. The GA works on a population of candidate solutions. First, the fitness of every individual in the population is evaluated in each iteration (called a ``generation''), where the fitness is the value of the objective function. The more fit individuals are stochastically selected from the current population and used to form a new generation by three biologically-inspired operators: mutation, crossover and selection. Finally, the algorithm terminates when a maximum number of generations or a satisfactory fitness level is reached.

In our Deep-ESN, we view the cascaded hyper-parameter vector of \textit{IS}, \textit{SR} and $\gamma$ of each reservoir as an individual, and the search space is constrained to the interval $[0,1]$. Additionally, we use the prediction error of the system as the fitness value of individual (the smaller loss, the higher fitness). We set a population size to 40 individuals and evaluate 80 generations. In all the experiments that follow, we use the training set to optimize the hyperparameters, with the fitness measured on the validation set.% {\LARGE GWC: I assume this last sentence is correct? (I added it).}

\subsection{Echo State Property in Deep-ESN}
\label{thoery:ESP}
The Echo State Property (ESP) is very important to ESNs, as it determines the stability and convergence of the whole system. In \cite{Jaeger2001The}, Jaeger gave a formal definition of the ESP:
	
\begin{definition}[\textbf{Echo-state property} \cite{Jaeger2001The}]
\emph{Assuming the standard compactness conditions (that is inputs $\mathbf{u}(t)$ and states $\mathbf{x}(t)$ come from a compact set $\mathscr{U}^{-\infty}$ and $\mathscr{X}^{-\infty}$ respectively) are satisfied. Moreover, assume the given ESN has no output feedback connections. Then the ESN can produce echo states (or has the ESP) if every echo-state $\mathbf{x}(t)$ is uniquely determined by every left-infinite input $\mathbf{u}^{-\infty}\in \mathscr{U}^{-\infty}$.} \label{def:ESP_ESN}	
\end{definition}

According to Definition.\ref{def:ESP_ESN}, we can see that nearby echo states have more similar input histories, which means that past information will be gradually washed out and the recent inputs and states are remembered. This is the so-called ``fading memory'' or ``short-term memory''. With this capacity, ESNs can accurately model the underlying dynamical characteristics of time series.

To ensure the stability of ESNs, Jaeger \cite{Jaeger2001The} also provided a sufficient condition for global asymptotic stability:

\begin{theorem}[\textbf{Global asymptotic condition} \cite{Jaeger2001The}]
	\emph{Let an ESN have fixed internal weight matrix $\mathbf{W}^{res}$ and the activation function in the reservoir is $f(\cdot)=tanh(\cdot)$, which satisfies the Lipschitz condition $\|f(x_1)-f(x_2)\|\le \|x_1-x_2\|$ for any $x_1,x_2\in\mathbb{R}$. Let $\mathbf{x}(t)$ and $\widetilde{\mathbf{x}}(t)$ be two distinct echo states at time $t$. If the largest singular value of the internal weight matrix $\overline{\sigma}(\mathbf{W}^{res})<1$, the ESN will have the echo-state property, i.e. , $\lim_{t\rightarrow {\infty}}{\|\mathbf{x}(t)-\widetilde{\mathbf{x}}(t)\|}\rightarrow 0$ for all right-infinite inputs $\mathbf{u}^{+\infty}\in \mathscr{U}^{+\infty}$.}
\label{thm:ESN}
\end{theorem}

Unlike the aforementioned necessary condition of stability ($SR<1$), the sufficient condition ($\overline{\sigma}(\mathbf{W}^{res})<1$) provides more restrictive theoretical support for global asymptotic stability.

In the following, we formally introduce the analysis of global asymptotic stability for our Deep-ESN.
For convenience, we consider leaky unit reservoir neurons, consistent with Jaeger's work \cite{Jaeger2001The}.

\begin{theorem}[\textbf{Stability condition for Deep-ESN}] \emph{Assume a Deep-ESN with $K$ reservoirs has fixed high-dimensional projection matrices $\mathbf{W}^{in(i)}$, internal matrix $\mathbf{W}^{res(i)}$ of each reservoir and the learned encoding matrix $\mathbf{W}^{enc(j)}$ where $\|\mathbf{W}^{in(i)}\|_2, \|\mathbf{W}^{enc(i-1)}\|_2$ both are bounded, $i=1,\dots,K$ and $j=1,\dots,K-1$. Let the activation function in reservoir is $f(\cdot)=tanh(\cdot)$, which satisfies the Lipschitz condition $\|f(x_1)-f(x_2)\|\le \|x_1-x_2\|$ for any $x_1,x_2\in\mathbb{R}$, and the encoders are linear for convenience. Moreover, we use $\mathbf{x}_{res}^{(i)}(t)$ and $\widetilde{\mathbf{x}}_{res}^{(i)}(t)$ to denote distinct echo states in $i$-th reservoir at time $t$. And then, if the largest singular value of internal weight matrix of each reservoir all satisfy the condition $\overline{\sigma}(\mathbf{W}^{res(i)})<1$, we say that the Deep-ESN will has the echo-state property, e.g., $\lim_{t\rightarrow {\infty}}{\|\mathbf{x}_{res}^{(i)}(t)-\widetilde{\mathbf{x}}_{res}^{(i)}(t)\|}\rightarrow 0$ for $i\in {1,\dots, K}$ and all right-infinite inputs $\mathbf{u}^{+\infty}\in \mathscr{U}^{+\infty}$.}
\end{theorem}

\begin{IEEEproof}
First, we consider the asymptotic condition in the first reservoir. At the time $t+1$, the input $\mathbf{u}(t+1)$ is projected by $\mathbf{W}^{in(1)}$. For two distinct echo states ($\mathbf{x}_{res}^{(1)}(t)$ and $\widetilde{\mathbf{x}}_{res}^{(1)}(t)$) at previous time $t$, we have their difference at the current time as follows:
\begin{align}
&\|\Delta\mathbf{x}^{(1)}_{res}(t+1)\|_2
=\|\mathbf{x}^{(1)}_{res}(t+1)\!-\!\widetilde{\mathbf{x}}^{(1)}_{res}(t+1)\|_2 \label{f1:1}\\
=&\|f(\mathbf{W}^{res(1)}\mathbf{x}_{res}^{(1)}(t)+\mathbf{W}^{in(1)}\mathbf{u}(t+1)) \\& -f(\mathbf{W}^{res(1)}\widetilde{\mathbf{x}}_{res}^{(1)}(t)+\mathbf{W}^{in(1)}\mathbf{u}(t+1))\|_2 \notag\\
\le&\|(\mathbf{W}^{res(1)}\mathbf{x}_{res}^{(1)}(t)+\mathbf{W}^{in(1)}\mathbf{u}(t+1)) \\& -(\mathbf{W}^{res(1)}\widetilde{\mathbf{x}}_{res}^{(1)}(t)+\mathbf{W}^{in(1)}\mathbf{u}(t+1))\|_2 \notag\\
=&\|\mathbf{W}^{res(1)}\mathbf{x}_{res}^{(1)}(t) -\mathbf{W}^{res(1)}\widetilde{\mathbf{x}}_{res}^{(1)}(t)\|_2 \notag\\
\le&\|\mathbf{W}^{res(1)}\|_2\|\Delta\mathbf{x}^{(1)}_{res}(t)\|_2\\
=&\overline{\sigma}(\mathbf{W}^{res(1)})\|\Delta\mathbf{x}^{(1)}_{res}(t)\|_2\label{f1:5}
\end{align}	
From the result of (\ref{f1:5}), $\lim_{t\rightarrow {\infty}}{\|\Delta\mathbf{x}^{(1)}_{res}(t+1)\|}\rightarrow 0$ is satisfied for all right-infinite inputs $\mathbf{u}^{+\infty}\in \mathscr{U}^{+\infty}$ when $\overline{\sigma}(\mathbf{W}^{res(1)})<1$. This result is consistent with that of a single-reservoir ESN.

For the following reservoirs ($i>1$), we have	
\begin{align}
&\|\Delta\mathbf{x}^{(i)}_{res}(t+1)\|_2
=\|\mathbf{x}^{(i)}_{res}(t+1)\!-\!\widetilde{\mathbf{x}}^{(i)}_{res}(t+1)\|_2 \\
=&\|f(\mathbf{W}^{res(i)}\mathbf{x}_{res}^{(i)}(t)+\mathbf{W}^{in(i)}\mathbf{x}_{enc}^{(i-1)}(t+1)) \\& -f(\mathbf{W}^{res(i)}\widetilde{\mathbf{x}}_{res}^{(i)}(t)+\mathbf{W}^{in(i)}\widetilde{\mathbf{x}}_{enc}^{(i-1)}(t+1))\|_2 \notag\\
\end{align}
\begin{align}
\le&\|(\mathbf{W}^{res(i)}\mathbf{x}_{res}^{(i)}(t)+\mathbf{W}^{in(i)}\mathbf{x}_{enc}^{(i-1)}(t+1)) \\&
-(\mathbf{W}^{res(i)}\widetilde{\mathbf{x}}_{res}^{(i)}(t)+\mathbf{W}^{in(i)}\widetilde{\mathbf{x}}_{enc}^{(i-1)}(t+1))\|_2 \notag\\
\le&\|\mathbf{W}^{res(i)}\|_2\|\mathbf{x}_{res}^{(i)}(t) -\widetilde{\mathbf{x}}_{res}^{(i)}(t))\|_2 \\&
+\|\mathbf{W}^{in(i)}(\mathbf{x}_{enc}^{(i-1)}(t+1)-\widetilde{\mathbf{x}}_{enc}^{(i-1)}(t+1))\|_2\notag\\
=&\|\mathbf{W}^{res(i)}\|_2\|\Delta\mathbf{x}^{(i)}_{res}(t)\|_2\\&
+\|\mathbf{W}^{in(i)}f(\mathbf{W}^{enc(i-1)}(\mathbf{x}_{res}^{(i-1)}(t+1)-\widetilde{\mathbf{x}}_{res}^{(i-1)}(t+1))\|_2\notag\\
\le&\|\mathbf{W}^{res(i)}\|_2\|\Delta\mathbf{x}^{(i)}_{res}(t)\|_2\\&
+\|\mathbf{W}^{in(i)}\mathbf{W}^{enc(i-1)}(\mathbf{x}_{res}^{(i-1)}(t+1)\!-\!\widetilde{\mathbf{x}}_{res}^{(i-1)}(t+1)\|_2\notag\\
%\end{align}
%\begin{align}
%\le&\|\mathbf{W}^{res(i)}\|_2\|\Delta\mathbf{x}^{(i)}_{res}(t)\|_2\\&
%+\|\mathbf{W}^{in(i)}\mathbf{W}^{enc(i-1)}\|_2\|\Delta\mathbf{x}^{(i-1)}_{res}(t+1)\|_2\notag\\
\le&\overline{\sigma}(\mathbf{W}^{res(i)})\|\Delta\mathbf{x}^{(i)}_{res}(t)\|_2\label{f2:8}\\&
+\|\mathbf{W}^{in(i)}\|_2\|\mathbf{W}^{enc(i-1)}\|_2\|\Delta\mathbf{x}^{(i-1)}_{res}(t+1)\|_2\notag
\end{align}

In the result of (\ref{f2:8}), the term $\|\Delta\mathbf{x}^{(i)}_{res}(t+1)\|_2$ is mainly bounded by $\|\Delta\mathbf{x}^{(i)}_{res}(t)\|_2$ and $\|\Delta\mathbf{x}^{(i-1)}_{res}(t+1)\|_2$ (two time-variant parts). Since the previous  results as $\lim_{t\rightarrow {\infty}}{\|\Delta\mathbf{x}^{(1)}_{res}(t+1)\|}\rightarrow 0$ with  $\overline{\sigma}(\mathbf{W}^{res(1)})<1$, we can have  $\lim_{t\rightarrow {\infty}}{\|\Delta\mathbf{x}^{(i)}_{res}(t+1)\|}\rightarrow 0$ by induction if it satisfies the condition $\overline{\sigma}(\mathbf{W}^{res(i')})<1$ for all $i'\in \{1,\dots,i\}$. In this situation, when $t\rightarrow {\infty}$, the first term in (\ref{f2:8}) will wash out with its factor $\overline{\sigma}(\mathbf{W}^{res(i')})<1$ and $\|\Delta\mathbf{x}^{(i-1)}_{res}(t+1)\|_2\rightarrow {0}$.
\end{IEEEproof}

\subsection{Computational Complexity}
\label{thoery:complexity}
Although Deep-ESN is a deep neural model of reservoir computing, there are not large additional costs in the whole learning process.
In this section, we analyze the computational complexity of Deep-ESN.

Assuming a Deep-ESN has $K$ reservoirs and $K-1$ PCA-based encoders, where sizes of the reservoirs are all fixed by $N$, and %and the sparseness of recurrent connections are $\alpha$ (that means $\mathbf{W}^{res}$ is a ($\alpha N^2$)-sparse matrix)
the reduced dimensionality is $M$ ($M<N$). Given $T$-length $D$-dimensional input sequences (assume the washout length of each reservoir \(T_{\mathrm{washout}}\) to be zero), we can analyze the computational complexity of Deep-ESN as follows.

For the high-dimensional projection (\ref{eq:project_drn}) and the update step (\ref{eq:update_drn}) in $i$-th reservoir, its complexity can be computed by
\begin{numcases}{\mathcal{C}_{res(i)}=}
\mathcal{O}(2\alpha TN^2+2TND),\ i=1, \\
\mathcal{O}(2\alpha TN^2+2TNM),\ i=2,3,\dots,K.
\end{numcases}
where $\alpha$ is very small (usually fixed by 0.01).

The complexity of encoding $j$-th states with PCA \cite{sharma-fast-pca} mentioned before can be computed by
%\begin{equation}{}
%\mathcal{C}_{enc(j)}=\mathcal{O}(\frac{1}{3}TMN),\ j=1, 2,\dots,K-1.
%\end{equation}
\begin{equation}{}
\mathcal{C}_{enc(j)}=\mathcal{O}(TN^2+N^2M),\ j=1, 2,\dots,K-1.
\end{equation}

After updating the echo states at all the time stamps and all the layers, we can collect the last reservoir states, inputs and all the middle-layer-encoded features into the $\mathbf{M}$ where its size is $(N+(K-1)M+D)\times T$ with full row rank. In this way, the complexity of solving the regression problem in (\ref{eq:train_weights}) can be computed by
\begin{align}{}
&\mathcal{C}_{regression}=\mathcal{O}((T+(P/3))P^2)\\
&where\ P\!=N\!+\!(K\!-\!1)M\!+\!D.
\end{align}
Since reservoir usually is a much larger hidden layer compared with inputs and encoders, we can assume that $N \gg M$ and $N \gg D$. And then, we have $P\approx N$. In this way, $\mathcal{C}_{regression}$ can be rewritten as about $\mathcal{O}(TN^2+N^3)$. Further, if $T$ is much larger than $N$ (high dimension property of time series), then we have $T \gg N$ and $\mathcal{C}_{regression}=\mathcal{O}(TN^2)$.

Thus in total, the computational complexity of Deep-ESN can be formulated by
\begin{align}{}
&\mathcal{C}_{Deep-ESN}=\sum_{i=1}^{K}\mathcal{C}_{res(i)}\!+\!\sum_{j=1}^{K\!-\!1}\mathcal{C}_{enc(j)}\!+\!\mathcal{C}_{regression}\\
&\!\approx\!\mathcal{O}(2\alpha TKN^2\!+\!2TND\!+\!(K\!-\!1)2TNM\!+\!(K\!-\!1)(TN^2\notag\\
&\!+\!N^2M)+TN^2)\!\approx\!\mathcal{O}(TN^2)
\label{eq:complexity}
\end{align}
It is seen that, with efficient unsupervised encoders (e.g., PCA), the computational complexity of Deep-ESN is $\mathcal{O}(TN^2)$. This is the $training$ complexity of Deep-ESN, and the run-time complexity is much smaller.

A single-reservoir ESN's computational complexity can be given by
\begin{align}{}
&\mathcal{C}_{ESN}=\mathcal{C}_{res}\!+\!\mathcal{C}_{regression}\\
&\!\approx\!\mathcal{O}(2\alpha TN^2\!+\!2TND\!+\!+TN^2)\!\approx\!\mathcal{O}(TN^2)
\label{eq:complexity_esn}
\end{align}

Therefore, Deep-ESN can realize equivalent computational performance to a single-reservoir ESN, which means that Deep-ESN remains the high computational efficiency of traditional reservoir computing networks.

%It is seen that, with efficient unsupervised encoders (e.g., PCA), the computational complexity of Deep-ESN is $\mathcal{O}(TN^2)$. Therefore, Deep-ESN can realize equivalent computational performance to a single-reservoir ESN, which means that Deep-ESN remains the high computational efficiency of traditional reservoir computing networks.

\section{Experiments}

\begin{figure}[!t]
	\centering
	\includegraphics[width=0.5\textwidth,height=1.7in]{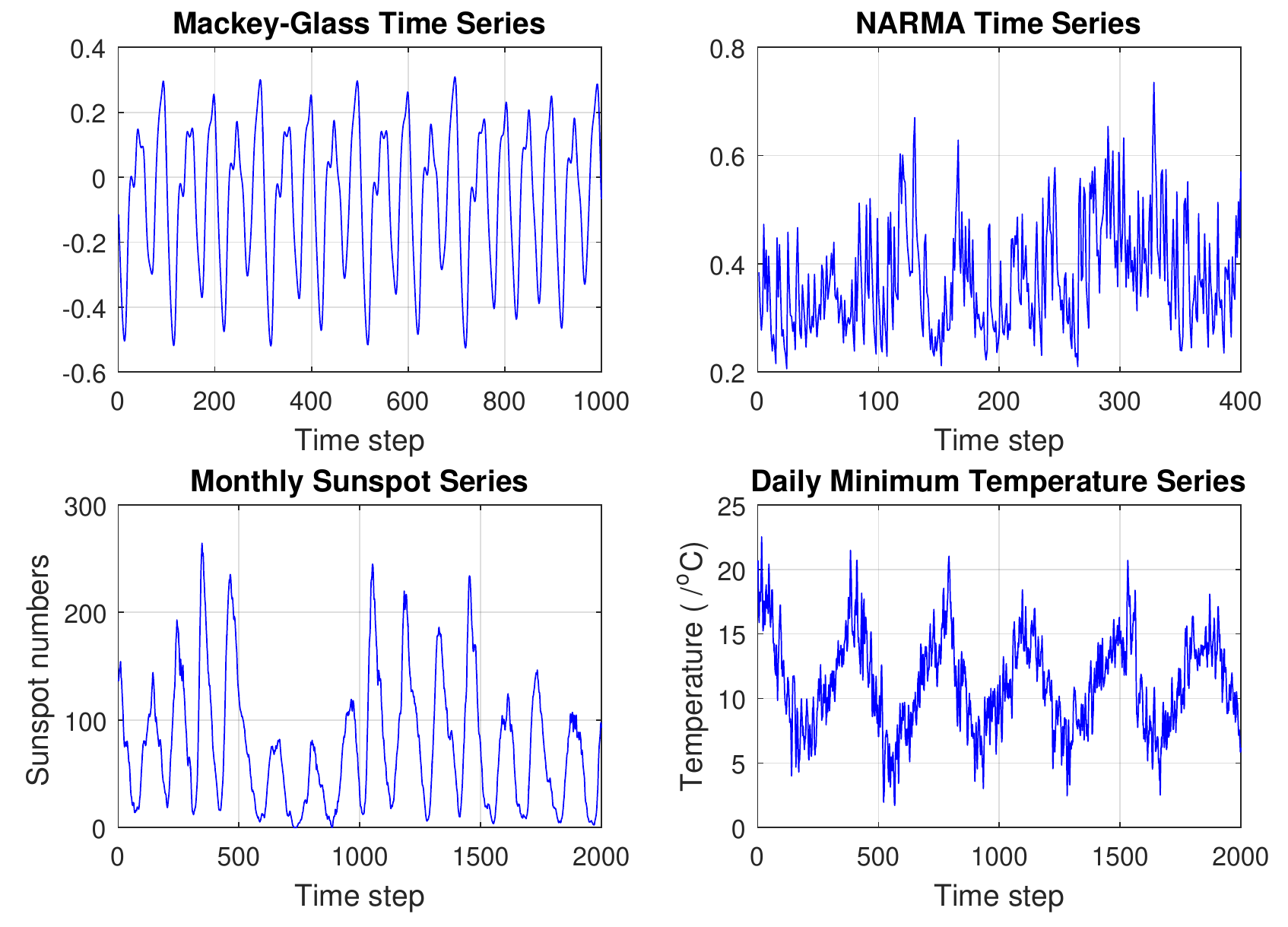}
	\caption{Visualization of selected time series, including chaotic time series (MGS and NARMA), real world time series (monthly sunspot series and daily minimum temperatures).}
	\label{data}
\end{figure}

In this section, we provide a comprehensive experimental analysis of our proposed Deep-ESN on two chaotic systems and two real world time series. Specifically, these time series are 1) the Mackey-Glass system (MGS); 2) NARMA system; 3) the monthly sunspot series and 4) a daily minimum-temperature series. Fig.~\ref{data} shows examples of these time series. Qualitatively, we see that the NARMA dataset and the daily minimum-temperature series present strong nonlinearity, the monthly sunspot series presents nonlinearity at its peaks, and the MGS chaotic series is relatively smooth.

To evaluate the effectiveness of proposed Deep-ESN, we compare with four baseline models, including a single-reservoir ESN with leaky neurons \cite{jaeger2007optimization}, the aforementioned two-layer ESN variants: $\varphi$-ESN \cite{gallicchio2011architectural}, R$^2$SP \cite{Butcher201376}, and the most recent hierarchical ESN-based model called multilayered echo-state machine (MESM) \cite{MESM}. In fact, MESM can be viewed as a simplified variant of the Deep-ESN without encoders. Since the core of our work is exploring and analyzing the effectiveness of  hierarchical schema, we ignore other variants of ESNs, for example, the Simple Cycle Reservoir (SCR) \cite{SCRESN} with a circular topology reservoir, and the support vector echo state machine (SVESM) \cite{SVESM2007} which optimizes the output weights with an SVM. It is worth noting that these single-reservoir variants (SCR, SVESM) can be viewed as a module and integrated into our Deep-ESN framework.

We also compare the performance among Deep-ESNs with various encoders (PCA, ELM-AE, RP). Furthermore, in these comparisons, we conduct experiments on Deep-ESNs with or without feature links to evaluate the impact of fusing multiscale dynamics to the outputs.

%{\Large GWC: What is the state of the art on these datasets? How do we compare to them?}
%Compared with the baseline MESM, our Deep-ESN has two main differences. One is using encoders as intermediate between two reservoirs, and the other one is the feature links from the encoders to output (seen in Fig.\ref{drn}).
%Therefore, we also design the corresponding control group to analyze their rationality.

The performance of all methods are evaluated by three widely used metrics: the root mean squared error (RMSE), the normalized root mean squared error (NRMSE) and the mean absolute percentage error (MAPE). They are used by most of ESN-based methods and can be formulated as follows.
\begin{align}
  &\mathrm{RMSE}=\sqrt{\frac{1}{T}\sum_{t=1}^{T}{[\mathbf{y}(t)-\hat{\mathbf{y}}(t)]^2}}\\
  &\mathrm{NRMSE}=\sqrt{{\sum_{t=1}^{T}{[\mathbf{y}(t)-\hat{\mathbf{y}}(t)]^2}}/\left\{\sum_{t=1}^{T}{[\mathbf{y}(t)-\overline{\mathbf{y}}]^2}\right\}}\\
  &\mathrm{MAPE}=\frac{1}{T}\sum_{t=1}^{T}{\frac{|\mathbf{y}(t)-\hat{\mathbf{y}}(t)|}{\mathbf{y}(t)}}\times100\%
\end{align}
%{\Large GWC: Since NRMSE is just a scaled version of RMSE, I'm not sure why you want to include both?}
where \(\mathbf{y}(t)\) denotes \(t\)-th observation from \(T\)-length target signals \(\mathbf{y}=[\mathbf{y}(1),\mathbf{y}(2),\dots,\mathbf{y}(T)]\). \(\overline{\mathbf{y}}\) denotes the mean of \(T\) observation points from \(\mathbf{y}\). \(\hat{\mathbf{y}}(t)\) denotes \(t\)-th observation output of the forecasting process. Note that the metric MAPE can not be used for evaluating the time series involving zero values (the denominator should not be zero), so we modify it by adding a small number, e.g., adding 0.1 to the sunspot time series and the predicted results before computing their MAPE.
For these three metrics, smaller values mean better performance.

In the following simulations, we adopt the metric RMSE to direct the GA-based hyperparameter optimization (hyperparameters: \(IS\), \(SR\) and \(\gamma\)). Although the baseline systems did not use the GA to choose their hyperparameters in the original work, here we use the GA to optimize all baselines to ensure fairness of comparison. For the MESM and Deep-ESNs, we report the best result among their variants with different number of layers (reservoirs) from 2 to 8 and different size of
encoders from 10 to 300 by intervals of 10 (N.B., a 2 layer Deep-ESN has two reservoirs and one encoder layer, while MESM only has two directly connected reservoirs). We use cross-validation to determine the optimal layer number and the encoder size. From our experiments, we find the Deep-ESNs with larger sizes of reservoirs have better performances. For simplicity, the sizes of reservoirs are fixed to 300. We will discuss the effects of the reservoir size in Section IV-E. More details of the parameters settings are given in Table~\ref{parms}.

\begin{table}[!t]
\caption{Parameter settings in our proposed Deep-ESN.}
\label{parms}
\renewcommand\arraystretch{1.3}
\centering
\begin{tabular}{l|c}
    \hline
    \multicolumn{2}{c}{GA search settings}\\
    \hline
    Input scaling $IS$ & [0, 1] \\
    Spectral radius $SR$ & [0, 1] \\
    Leaky rate $\gamma$ & [0, 1] \\
    Population size & 40 \\
    Generations     & 80 \\
    \hline
    \hline
    \multicolumn{2}{c}{Fixed hyper-parameters}\\
    \hline
    Sparsity $\alpha$ & 10\% \\
    Size of each reservoir $N$ & 300 \\
    Activation function of reservoir $f(\cdot)$ & $tanh(\cdot)$ \\
\hline
\end{tabular}
\end{table}

\subsection{Mackey-Glass System}
\label{sec:mgs}

The Mackey-Glass System (MGS) is a classical time series for evaluating the performance of dynamical system identification methods \cite{Jaeger2004Harnessing,Jaeger2001The}. In discrete time, the Mackey-Glass delay differential equation can be formulated by
\begin{align}
  &y(t+1)=y(t)+\delta\cdot(a\frac{y(t-\tau/\delta)}{1+y(t-\tau/\delta)^{n}}-by(t))
\label{eq:MGS}
\end{align}
where the parameters \(\delta\), \(a\), \(b\), \(n\) usually are set to 0.1, 0.2, -0.1, 10. When \(\tau>16.8\), the system becomes chaotic, and in most previous work, \(\tau\) is set to 17. Thus, we also let \(\tau\) be 17 in this task. In detail, we generate a 10000-length time series from Equation (\ref{eq:MGS}) by the fourth-order Runge-Kutta method. We split these 10000 points into three parts with length \(T_{\mathrm{train}}=6400\), \(T_{\mathrm{validate}}=1600\) and \(T_{\mathrm{test}}=2000\). To avoid the influence of initial states, we discard a certain number of initial steps,  \(T_{\mathrm{washout}}=100\) for each reservoir.

The task is to predict the input 84 time steps ahead. Thus, all of the methods are required to learn the mapping from the current input $\mathbf{u}(t)$ to the target output $\mathbf{y}(t)$, where $\mathbf{y}(t)=\mathbf{u}(t+84)$. For each model, we conduct ten simulations independently and record their the average result and standard deviation.

The detailed results are presented in Table \ref{results:84mgs}. As seen in the Table, all of the Deep-ESN models (with feature links) outperform the baselines by about an order of magnitude.  Among all of the baseline systems, $\varphi$-ESN \cite{gallicchio2011architectural} is the best. MESM reaches its best result with 7 reservoirs.

\begin{table*}[t!]
\caption{Average Results with Standard Deviations of 84-Step-Ahead Prediction for Mackey-Glass System}
\label{results:84mgs}
\renewcommand\arraystretch{1.3}
\scriptsize
\centering
\begin{tabular}{l|c|ccc|c|c}
    \hline
    Methods & Feature links& RMSE & NRMSE & MAPE & $\sharp$ Layers & Size of Encoders\\
    \hline\hline
    ESN \cite{jaeger2007optimization}	& - & 4.37E-02 $\pm$ 6.31E-03 & 2.01E-01 $\pm$ 2.91E-02 & 7.03E-01 $\pm$ 1.27E-01 & 1&-\\
    $\varphi$-ESN \cite{gallicchio2011architectural}& -  & 8.60E-03 $\pm$ 1.63E-03 & 3.96E-02 $\pm$ 7.49E-03 & 1.00E-01 $\pm$ 2.13E-02& 2&-\\
    R$^2$SP \cite{Butcher201376}& -  & 2.72E-02 $\pm$ 4.27E-03 & 1.25E-01 $\pm$ 1.96E-02 & 1.00E-01 $\pm$ 2.13E-02& 2&-\\
    MESM \cite{MESM} & - & 1.27E-02 $\pm$ 2.50E-03 & 5.86E-02 $\pm$ 1.15E-02 & 1.91E-01 $\pm$ 4.22E-02 & 7&-\\
    \hline
    \multirow{2}{3cm}{Deep-ESN with PCA}
     & yes  & $\textbf{1.12E-03} \pm \textbf{1.87E-04}$ & $\textbf{5.17E-03} \pm \textbf{8.61E-04}$ & $\textbf{1.51E-02} \pm \textbf{3.06E-03}$& \multirow{2}{0.5cm}{\centering 8} &\multirow{2}{1.3cm}{\centering 110}\\
     & no   & 3.25E-03 $\pm$ 4.54E-04 & 1.49E-02 $\pm$ 2.09E-03 & 3.89E-02 $\pm$ 7.41E-03 & &\\
    \hline
    \multirow{2}{3cm}{Deep-ESN with ELM-AE}
    & yes & 1.49E-03 $\pm$ 2.98E-04 & 6.84E-03 $\pm$ 1.37E-03 & 1.56E-02 $\pm$ 4.02E-03& \multirow{2}{0.5cm}{\centering 8} &\multirow{2}{1.3cm}{\centering 140}\\
    & no  & 2.99E-03 $\pm$ 4.23E-04 & 1.38E-02 $\pm$ 1.94E-03 & 3.72E-02 $\pm$ 8.52E-03&  & \\
    \hline
    \multirow{2}{3cm}{Deep-ESN with RP}
    &  yes  & 1.57E-03 $\pm$ 9.49E-04 & 7.20E-03 $\pm$ 4.36E-03 & 1.86E-02 $\pm$ 1.13E-02& \multirow{2}{0.5cm}{\centering 8} &\multirow{2}{1.3cm}{\centering 150}\\
    &  no  & 4.29E-03 $\pm$ 2.02E-03 & 1.97E-02 $\pm$ 9.31E-03 & 5.40E-02 $\pm$ 2.24E-02 &  & \\
\hline
\end{tabular}
\end{table*}

\begin{figure}[t!]
\centering
\begin{minipage}{0.45\linewidth}
\centering
  \includegraphics[width=0.98\textwidth]{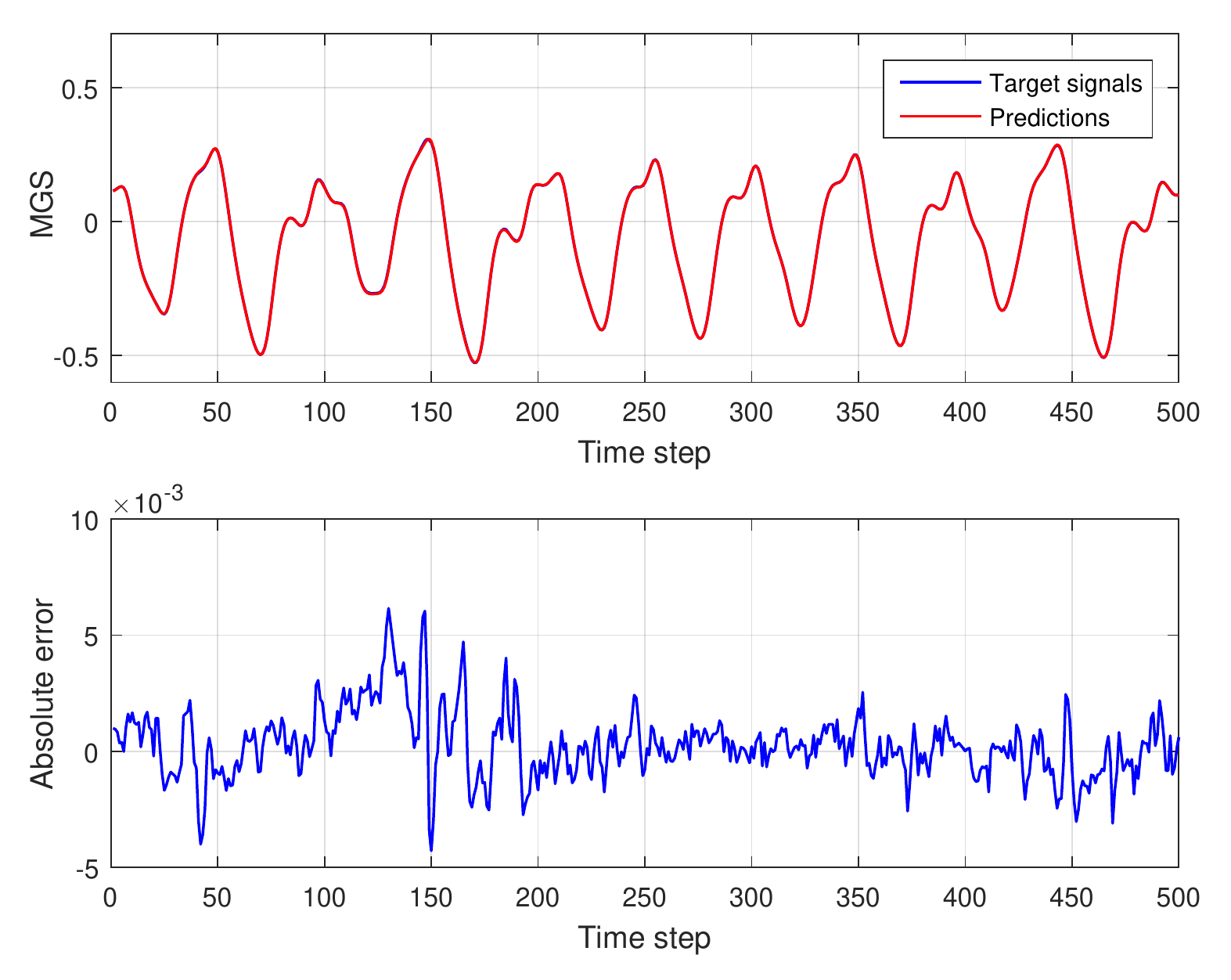}
  \centerline{(a)}
  \label{fig:aberror_mgs}
\end{minipage}
% \hfill
\begin{minipage}{.45\linewidth}
\centering
  \includegraphics[width=0.98\textwidth]{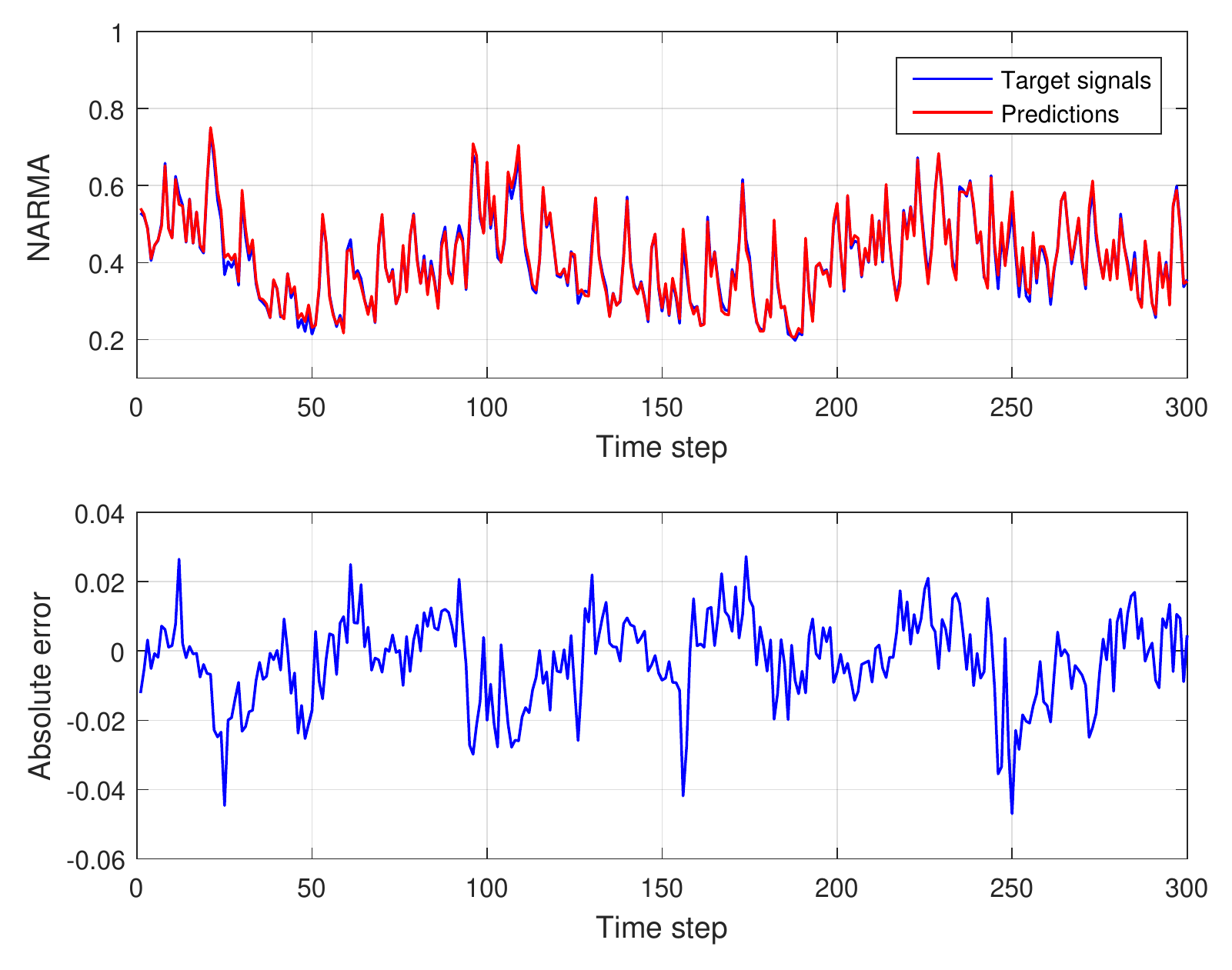}
  \centerline{(b)}
  \label{fig:aberror_narma}
\end{minipage}
\begin{minipage}{.45\linewidth}
\centering
  \includegraphics[width=0.98\textwidth]{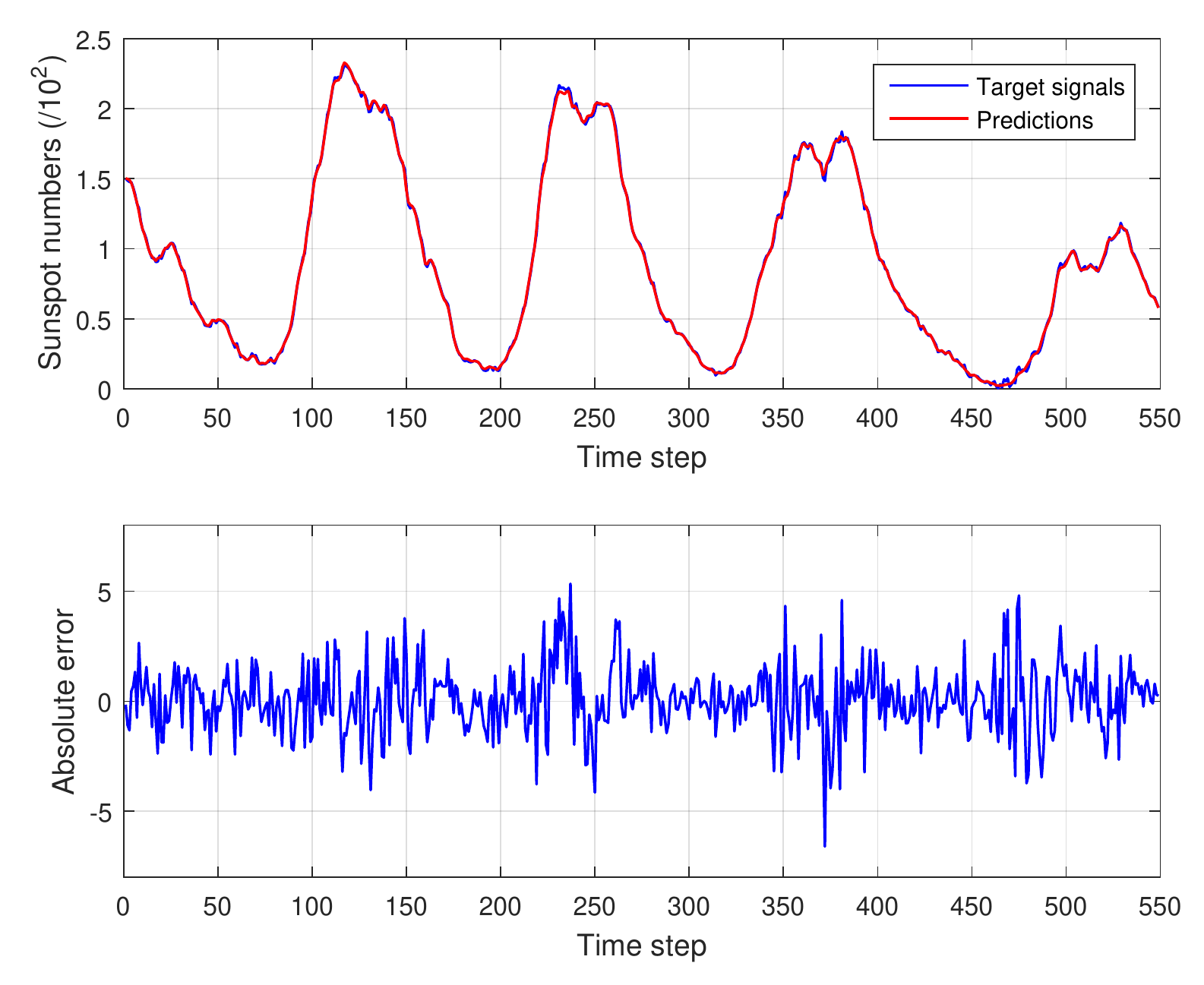}
  \centerline{(c)}
  \label{fig:aberror_sunspot}
\end{minipage}
\begin{minipage}{.45\linewidth}
\centering
  \includegraphics[width=0.98\textwidth]{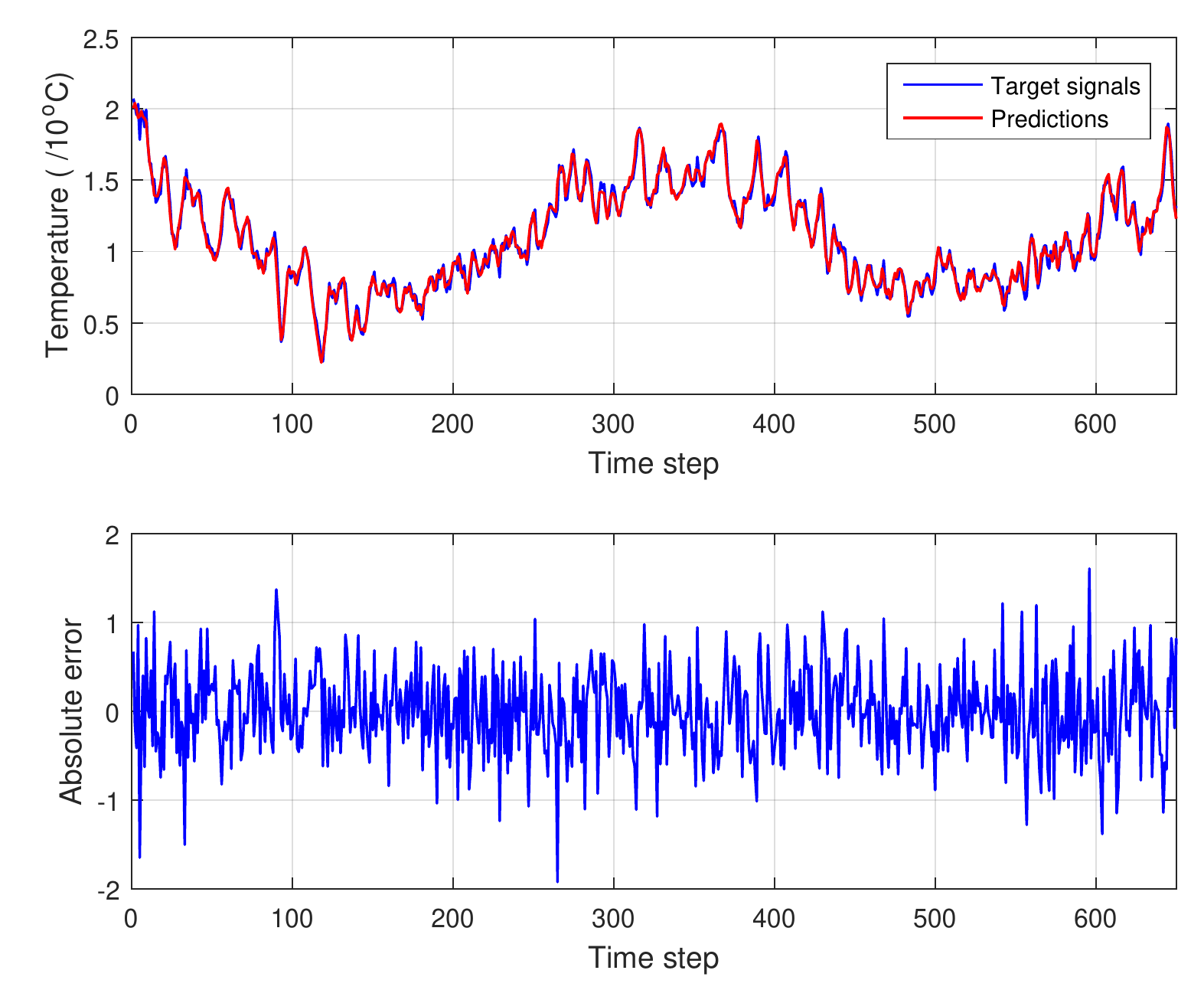}
  \centerline{(d)}
  \label{fig:aberror_temperature}
\end{minipage}
\caption{Prediction curves and absolute error. All results are for the Deep-ESN using PCA for the encoder. (a) 84-step-ahead prediction over Mackey-Glass time series; (b) one-step-ahead prediction over NARMA time series; (c) one-step prediction over Monthly Sunspot Number time series; (d) one-step prediction over Daily Minimum Temperature time series.}
\label{fig:aberror12}
\end{figure}

\begin{table*}[t!]	
\caption{Average Results with Standard Deviations of One-Step-Ahead Prediction for Tenth-Order NARMA}
\label{results:narma}
\renewcommand\arraystretch{1.3}
\centering
\scriptsize
\begin{tabular}{l|c|ccc|c|c}
\hline
Methods & Feature links& RMSE & NRMSE & MAPE & $\sharp$ Layers& Size of Encoders\\
\hline\hline
ESN \cite{jaeger2007optimization}&-	& 2.76E-02 $\pm$ 2.25E-03 & 2.45E-01 $\pm$ 2.00E-02 & 5.72E-02 $\pm$ 5.01E-03 & 1 &-\\
$\varphi$-ESN \cite{gallicchio2011architectural}&- & 1.92E-02 $\pm$ 2.00E-03 & 1.69E-01 $\pm$ 1.75E-02 & 3.94E-02 $\pm$ 4.13E-03 & 2 &-\\
R$^2$SP \cite{Butcher201376}&- & 2.05E-02 $\pm$ 2.38E-03 & 1.81E-01 $\pm$ 2.21E-02 & 4.30E-02 $\pm$ 5.43E-03 & 2 &-\\
MESM \cite{MESM}&- & 1.91E-02 $\pm$ 2.73E-03 & 1.68E-01 $\pm$ 2.40E-02 & 4.07E-02 $\pm$ 5.59E-03 & 2 &-\\
\hline
\multirow{2}{3cm}{Deep-ESN with PCA}
&yes   & $\textbf{1.39E-02} \pm \textbf{\textbf{1.33E-03}}$ & $\textbf{1.21E-01} \pm \textbf{1.16E-02}$ & $\textbf{2.19E-02} \pm \textbf{2.48E-03}$ & \multirow{2}{0.5cm}{\centering 4} &\multirow{2}{1.3cm}{\centering 280} \\
&no    & 2.37E-02 $\pm$ 2.87E-03 & 2.06E-01 $\pm$ 2.50E-02 & 3.91E-02 $\pm$ 4.91E-03 &  & \\
\hline
\multirow{2}{3cm}{Deep-ESN with ELM-AE}
&yes & 1.55E-02 $\pm$ 1.49E-03 & 1.36E-01 $\pm$ 1.30E-02 & 2.51E-02 $\pm$ 2.40E-03 & \multirow{2}{0.5cm}{\centering 3} &\multirow{2}{1.3cm}{\centering 70} \\
&no  & 1.85E-02 $\pm$ 1.64E-03 & 1.61E-01 $\pm$ 1.43E-02 & 3.00E-02 $\pm$ 2.82E-03 & &\\
\hline
\multirow{2}{3cm}{Deep-ESN with RP}  	
&yes & 1.43E-02 $\pm$ 1.75E-03 & 1.25E-01 $\pm$ 1.54E-02 & 2.30E-02 $\pm$ 2.94E-03 & \multirow{2}{0.5cm}{\centering 2} &\multirow{2}{1.3cm}{\centering 280} \\
&no  & 1.60E-02 $\pm$ 1.40E-03 & 1.41E-01 $\pm$ 1.23E-02 & 2.61E-02 $\pm$ 2.36E-03 & &\\
\hline
\end{tabular}
\end{table*}

For the case of Deep-ESN with different encoders, the PCA variant outperforms the ELM-AE and RP encoders in this task, a result that will be reprised in every experiment. We also found that the depth of the network has a significant effect on the Deep-ESN performance. We tried up to 8 layers, and performance improved as we added layers. Further analysis of the effects of depth on network performance is in Section \ref{subsec:network_structure}.

In all models, removing the feature links reduces performance. Again, this will be a theme that will be repeated in the remaining experiments. This demonstrates that incorporating the feature links are a crucial part of the design of the Deep-ESN, as they provide multiscale information to the output layer, compared to networks where the output only depends on the dynamics of the last reservoir.

Fig.\ref{fig:aberror12} (a) presents the prediction curve and absolute error curve of MGS time series simulated by Deep-ESN in the testing phase. It should be noted that the target curve is obscured by the system output, so the Deep-ESN fits the chaotic time series very well with small errors.

\subsection{Tenth-Order NARMA System}
NARMA is short for nonlinear autoregressive moving average system, a highly nonlinear system incorporating memory. The tenth-order NARMA depends on outputs and inputs 9 time steps back, so it is considered difficult to identify. The tenth-order NARMA system can be described by
\begin{align}
	y(t+1)&=0.3\cdot y(t)+0.05\cdot y(t)\cdot \sum_{i=0}^9{y(t-i)}\notag\\
	&+1.5\cdot u(t-9)\cdot u(t)+0.1
	\label{eq:NARMA}
\end{align}
where $u(t)$ is a random input at time step $t$, drawn from a uniform distribution over [0, 0.5]. The output signal $y(t)$ is initialized by zeros for the first ten steps ($t=1,2,\dots,10$).  In this task, we generate a NARMA time series with total length \(T_{\mathrm{total}}=4000\) and split it into three parts with length \(T_{\mathrm{train}}=2560\), \(T_{\mathrm{validate}}=640\) and \(T_{\mathrm{test}}=800\), respectively. The washout length is also set to be 30 for each reservoir in all algorithms. We conduct one-step-ahead prediction of the tenth-order NARMA time series.

The averaged results of 10 independent simulations are presented in Table \ref{results:narma}. Again, all Deep-ESN models outperform the baselines, and the PCA encoder version is best. Again, the results of the Deep-ESNs with feature links outperform the ones without these links. MESM is the best among the baselines. We note that, compared with the results on the Mackey-Glass system, the hierarchical methods (MESM and Deep-ESNs) with the best performance tend to use fewer layers on the NARMA task.

Fig.\ref{fig:aberror12} (b) illustrates the prediction curve and absolute error curve of the 4-layer Deep-ESN with PCA on the 10th-order NARMA time series. In this case, errors are visible in the plot due to some strong nonlinear changes, which are mirrored in the error plot.

\subsection{Monthly Sunspot Series Forecasting}
Sunspot number is a dynamic manifestation of the strong magnetic field in the Sun's outer regions. It has been found that the sunspot number has a very close statistical relationship with solar activity \cite{Bray1979Sunspots}. Due to the complexity of the underlying solar activity and the high nonlinearity of the sunspot series, forecasting and analyzing these series is a challenging task, and is commonly used to evaluate the capability of time-series prediction models~\cite{Suyal2009,GrowingESN}. An open source 13-month smoothed monthly sunspot series is provided by the world data center SILSO \cite{sunspot}. We use the data from July, 1749 to November, 2016 in our forecasting experiment. There are a total of 3209 sample points. Since the last 11 points are still provisional and are subject to  possible revision, we remove these points and use the remaining 3198 observations. We split the data into three parts with length \(T_{\mathrm{train}}=2046\), \(T_{\mathrm{validate}}=512\) and \(T_{\mathrm{test}}=640\),  respectively. The washout length \(T_{\mathrm{washout}}\) for each reservoir is set to 30.

In this task, one-step-ahead prediction of sunspot time series is conducted ten times independently. The average results and standard deviations are reported in Table \ref{results:sunspot}. We can see that in this real world time series, our proposed Deep-ESNs also outperform the baselines. Under the metrics RMSE and NRMSE, the 3-layer Deep-ESN with PCA achieves the best performance. In an exception to previous results, here the Deep-ESN with ELM-AE presents the best result with MAPE. The feature links improve performance over not having feature links.

For the absolute error plot in Fig.\ref{fig:aberror12} (c), we see that the predicted results are in accord with the target signals on the whole, with  occasional error spikes on a minority of time points, e.g., at around 370 and 470 time steps.

\begin{table*}[t!]
\caption{Average Results with Standard Deviations of One-Step-Ahead Prediction for Monthly Sunspot Number Series}
\label{results:sunspot}
\renewcommand\arraystretch{1.3}
\centering
\scriptsize
\begin{tabular}{l|c|ccc|c|c}
\hline
Methods & Feature links & RMSE & NRMSE & MAPE & $\sharp$ Layers& Size of Encoders\\
\hline
ESN \cite{jaeger2007optimization}&-	& 1.30E-03 $\pm$ 7.43E-06 & 2.08E-02 $\pm$ 1.16E-04 & 4.96E-03 $\pm$ 8.74E-06 & 1 &-\\
$\varphi$-ESN \cite{gallicchio2011architectural}&-	 & 1.25E-03 $\pm$ 2.28E-05 & 1.93E-02 $\pm$ 3.51E-04 & 4.77E-03 $\pm$ 4.00E-05 & 2&- \\
R$^2$SP \cite{Butcher201376}&-	 & 1.27E-03 $\pm$ 2.44E-05 & 1.98E-02 $\pm$ 3.81E-04 & 4.98E-03 $\pm$ 1.23E-04 & 2&- \\
MESM \cite{MESM}&-	 & 1.26E-03 $\pm$ 3.08E-05 & 1.94E-02 $\pm$ 4.72E-04 & 4.87E-03 $\pm$ 9.15E-05 & 3&- \\
\hline
\multirow{2}{3cm}{Deep-ESN with PCA}
&yes & $\textbf{1.22E-03}\pm \textbf{1.24E-09}$ & $\textbf{1.87E-02} \pm \textbf{1.89E-04}$ & 4.76E-03 $\pm$ 2.83E-05 & \multirow{2}{0.5cm}{\centering 3} &\multirow{2}{1.3cm}{\centering 10}\\
&no  & 1.24E-03 $\pm$ 2.83E-05&1.90E-02 $\pm$ 4.43E-04& 4.82E-03 $\pm$ 1.11E-04 &&\\
\hline
\multirow{2}{3cm}{Deep-ESN with ELM-AE}
&yes & 1.23E-03 $\pm$ 2.44E-05 & 1.89E-02 $\pm$ 3.76E-04 & $\textbf{4.67E-03} \pm \textbf{6.02E-05}$ & \multirow{2}{0.5cm}{\centering 2} &\multirow{2}{1.3cm}{\centering 10} \\
&no  & 1.23E-03 $\pm$ 2.55E-05 & 1.89E-02 $\pm$ 3.93E-04 & 4.68E-03 $\pm$ 8.73E-05 &&\\
\hline
\multirow{2}{3cm}{Deep-ESN with RP}
	&yes & 1.24E-03 $\pm$ 3.10E-05 & 1.90E-02 $\pm$ 4.75E-04 & 4.79E-03 $\pm$ 8.46E-05 & \multirow{2}{0.5cm}{\centering 3} &\multirow{2}{1.3cm}{\centering 270} \\
    &no  & 1.26E-03 $\pm$ 2.07E-05 & 1.93E-02 $\pm$ 3.17E-04 & 4.82E-03 $\pm$ 4.47E-05 & &\\
\hline
\end{tabular}
\end{table*}

\begin{table*}[t!]
\caption{Average Results with Standard Deviations of One-Step-Ahead Prediction for Daily Minimum Temperatures in Melbourne, Australia}
\label{results:temperature}
\renewcommand\arraystretch{1.3}
\centering
\scriptsize
\begin{tabular}{l|c|ccc|c|c}
\hline
Methods  & Feature links & RMSE & NRMSE & MAPE & $\sharp$ Layers& Size of Encoders\\
\hline
ESN \cite{jaeger2007optimization} &- & 5.01E-01 $\pm$ 3.70E-03 & 1.39E-01 $\pm$ 1.02E-03 & 3.95E-02 $\pm$ 2.37E-04 & 1 &-\\
$\varphi$-ESN \cite{gallicchio2011architectural} &-  & 4.93E-01 $\pm$ 3.86E-03 & 1.41E-01 $\pm$ 1.10E-03 & 3.96E-02 $\pm$ 3.74E-04 & 2 &-\\
R$^2$SP \cite{Butcher201376}&-  & 4.95E-01 $\pm$ 3.55E-03 & 1.37E-01 $\pm$ 9.82E-04 & 3.93E-02 $\pm$ 4.34E-04 & 2&- \\
MESM \cite{MESM}&-  & 4.78E-01 $\pm$ 3.39E-03 & 1.36E-01 $\pm$ 9.67E-04 & 3.77E-02 $\pm$ 3.36E-04 & 2 &-\\
\hline
\multirow{2}{3cm}{Deep-ESN with PCA}
&yes  & $\textbf{4.73E-01}\pm \textbf{2.77E-03}$ & $\textbf{1.35E-01}\pm \textbf{7.91E-04}$ & $\textbf{3.70E-02}\pm \textbf{2.14E-04}$ & \multirow{2}{0.5cm}{\centering 2} &\multirow{2}{1.3cm}{\centering 240} \\
&no  &4.74E-01 $\pm$ 1.14E-03&1.35E-01 $\pm$ 4.15E-04&3.71E-02 $\pm$ 2.07E-04&&\\
\hline
\multirow{2}{3cm}{Deep-ESN with ELM-AE}
&yes & 4.75E-01 $\pm$ 2.27E-03 & 1.36E-01 $\pm$ 6.48E-04 & 3.73E-02 $\pm$ 2.36E-04 & \multirow{2}{0.5cm}{\centering 2} &\multirow{2}{1.3cm}{\centering 40} \\
&no  & 4.78E-01 $\pm$ 5.32E-03 & 1.36E-01 $\pm$ 1.52E-03 & 3.75E-02 $\pm$ 5.85E-04 & & \\
\hline
\multirow{2}{3cm}{Deep-ESN with RP}  	
&yes & 4.76E-01 $\pm$ 4.97E-03 & 1.36E-01 $\pm$ 1.42E-03 & 3.76E-02 $\pm$ 4.73E-04 & \multirow{2}{0.5cm}{\centering 2} &\multirow{2}{1.3cm}{\centering 240} \\
&no  & 4.77E-01 $\pm$ 2.70E-03 & 1.36E-01 $\pm$ 7.71E-04 & 3.75E-02 $\pm$ 3.05E-04 & &\\
\hline
\end{tabular}
\end{table*}

\subsection{Daily Minimum Temperatures Prediction}
\label{sec:temperature}
Daily minimum temperatures in Melbourne, Australia is our second real world time series dataset, recorded from January 1, 1981 to December 31, 1990 \cite{temperature}. There are a total of 3650 sample points (\(T_{\mathrm{total}}=3650\)). We let \(T_{\mathrm{train}}=2336\), \(T_{\mathrm{validate}}=584\) and \(T_{\mathrm{test}}=730\), respectively. The washout length \(T_{\mathrm{washout}}\) for each reservoir is set to 30. Since this real world time series shows strong nonlinearity, we smooth it with a 5-step sliding window. The smoothed data can be seen in Fig.\ref{data}.

We perform one-step-ahead prediction on the daily minimum temperatures. The results are listed in Table \ref{results:temperature}. As with our previous observations, the Deep-ESN with PCA gives the best results (with two-layer structure on this task). MESM also performs well and is better than ESN, $\varphi$-ESN and R$^2$SP with its two-layer reservoirs. Here, the feature links do not make much of a difference, suggesting that the time series is not multiscale. 
%Even so, the feature links still generally improve performance slightly over not having feature
%links. %networks with and without feature links perform nearly identically, which may be due to the relatively low number of layers.

Fig.\ref{fig:aberror12} (d) shows the predicted results and absolute errors plots of the test data. From these plots, we can see that the error signals show drastic oscillations due to the strong nonlinearity of the target time series, reflecting the high complexity of this dynamical system.

Over these four experiments, the Mackey-Glass 84-step-ahead prediction task needs the deepest structure (8 reservoirs), while one-step-ahead prediction of the daily minimum temperature series required shallower networks (2 reservoirs). We believe that this reflects the four time series have different multiscale structures.
%the amount of memory required for the different tasks, and that the ones with high nonlinearity may not need the same amount of memory. 
Furthermore, it is interesting to note that the feature links have much larger effects in deeper models (e.g. Deep-ESN with 8 reservoirs) than in shallower ones, which suggests that the time series with shallower models are not multiscale. Based on these informal observations, we provide a more detailed analysis of the effects of network structure in the next section.

\subsection{Analysis of Network Structure}
\label{subsec:network_structure}

In this section, we exam the effects of various structure parameters on the performance of 8-layer Deep-ESNs.

\begin{figure*}[t!]
\centering
\begin{minipage}{0.44\linewidth}
\centering
  \includegraphics[width=0.92\textwidth,height=1.7in]{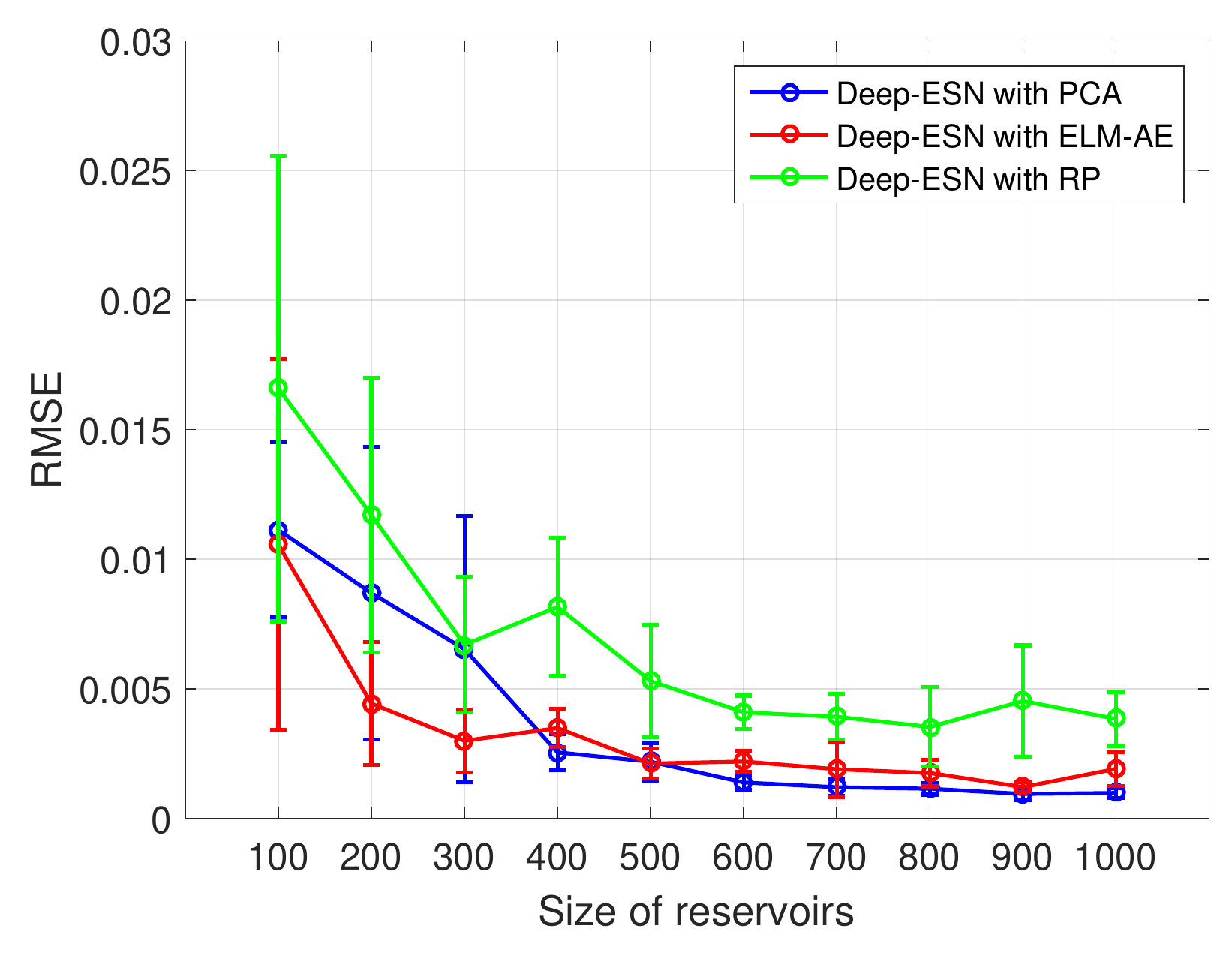}
  \centerline{(a)}
\label{analysis_res_size}
\end{minipage}
\begin{minipage}{0.44\linewidth}
\centering
  \includegraphics[width=0.92\textwidth,height=1.7in]{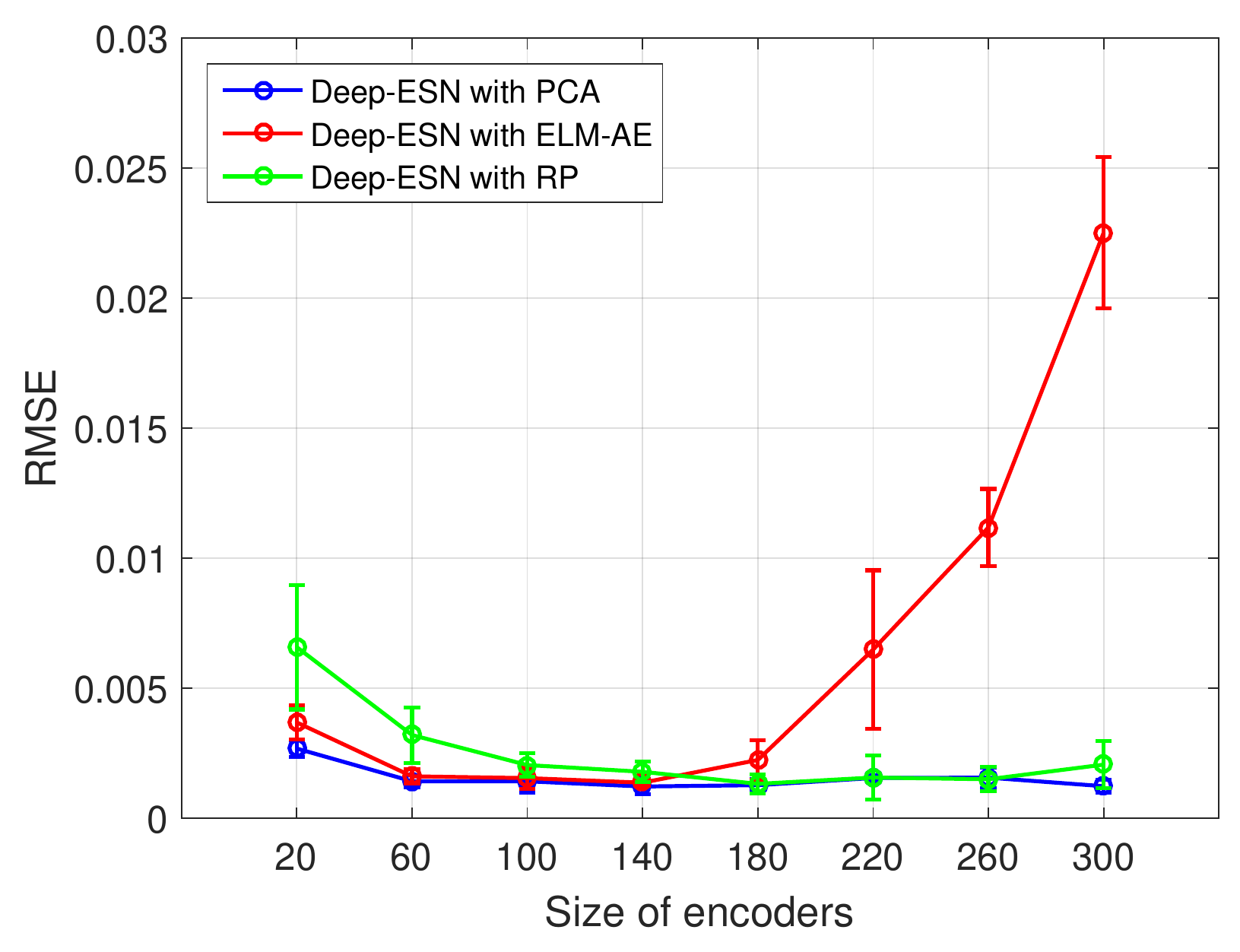}
  \centerline{(b)}
\label{analysis_enc_size}
\end{minipage}
%\hfill
%\end{tabular}
\centering
\caption{Investigation of the effects of the size of reservoirs and encoders in an 8-layer Deep-ESN. (a) Fixing the size of each encoder at 30 and increasing the size of all the reservoirs from 100 to 1000. (b) Fixing the size of all reservoirs at 300, and increasing the size (dimensionality) of each encoder from 20 to 300.}
\label{analysis_size}
\end{figure*}

%\begin{figure}[!t]
%	\centering
%	\includegraphics[width=0.48\textwidth,height=2.0in]{figures/fig1_change_res}
%	\caption{Analysis of the size of reservoir.}
%	\label{analysis_res_size}
%\end{figure}
%
%\begin{figure}[!t]
%	\centering
%	\includegraphics[width=0.48\textwidth,height=2.0in]{figures/fig2_change_encoder}
%	\caption{Analysis of the size of encoder.}
%	\label{analysis_enc_size}
%\end{figure}

First, we examine the effects of the size $N$ of the reservoir on the 84-step-ahead prediction of MGS time series. In this experiment, we used an 8-layer Deep-ESN with PCA. As illustrated in Fig.\ref{analysis_size} (a), we fix the size of each encoder at 30 and increase the size of all the reservoirs from 100 to 1000. %2) fixing the last (8th) reservoir at 100, and enlarging the other reservoirs from 100 to 1000 (red); 3) only enlarging the last reservoir, but fixing the size of the other reservoirs to 100 (green).
We can see that enlarging reservoirs generally improves the overall performance. This observation shows that the size of reservoir plays an important role in Deep-ESN. Deep-ESN with PCA has relatively better performance than the ones with ELM-AE or RP.
%Increasing the size of all of the reservoirs performs best and achieves the lowest error (RMSE).

Second, we look at the effects of encoder dimensionality in Figure~\ref{analysis_size}(b). We fix the size of all reservoirs at 300, and increase the encoder dimension from 20 to 300 by intervals of 40. As seen in Figure~\ref{analysis_size}(b), the dimensionality of the encoders also affects the performance of Deep-ESNs to some extent. If the encoder dimension is too small, it will lose too much information from the previous reservoir. Generally, a Deep-ESN with larger encoders performs better than one with smaller ones, with diminishing returns after about 60 dimensions for this problem. The effect persists with a flat slope, and we suspect this effect is due to the orthogonalization of the variables. If we increase the dimensionality of the encoder to match the reservoir (300D), it is interest to see that Deep-ESNs with PCA or RP still have good performances, while the performances of Deep-ESNs with ELM-AE continue to get rapidly worse when the encoder size is larger than 180. Therefore, it is better to use smaller size for ELM-AE.
%{\LARGE GWC: Here, you should explain how you chose the number of dimensions for your experiments,, if you haven't already (actually, this should be explained way back in the beginning when you explain how you used the GA to optimize your meta-parameters. I assume you used cross-validation for this.}
%In this task, for example, when the size of encoders is increasing from 20 to 60, we have lower errors of Deep-ESN with PCA. When the size is over 60, its performance will not improve significantly.

\begin{table}[!t]
\caption{Parameter settings in Fig.\ref{auto_add_layers}. }
\label{parms_analysis}
\renewcommand\arraystretch{1.3}
\centering
\begin{tabular}{lcccc}
    \hline
    model& indices of layer & $IS$ & $SR$ & $\gamma$ \\
    \hline
    \hline
    \multirow{3}{*}{Deep-ESN} & 1st & 0.7726 & 0.8896  & 0.2618\\
                              & 2nd & 0.4788 & 0.8948  & 0.6311\\
                              & 3rd & 0.6535 & 0.3782  & 0.2868\\
    \hline
    \multirow{3}{*}{MESM}     & 1st & 0.8054 & 0.8353  & 0.2916\\
                              & 2nd & 0.9526 & 0.9526  & 0.9695\\
                              & 3rd & 0.9903 & 0.9764  & 0.1893\\
    \hline
    \multicolumn{5}{l}{Other hyper-parameters}\\
    \hline
    \hline
    \multicolumn{3}{l}{Sparsity $\alpha$ }& \multicolumn{2}{c}{10\%} \\
    \multicolumn{3}{l}{Size of each reservoir $N$ }& \multicolumn{2}{c}{300} \\
    \multicolumn{3}{l}{Size of each encoder $M$ }& \multicolumn{2}{c}{30} \\
    \multicolumn{3}{l}{Activation function of reservoir $f(\cdot)$ }& \multicolumn{2}{c}{$tanh(\cdot)$} \\
\hline
\end{tabular}
\end{table}

\begin{figure}[t!]
\centering
\begin{minipage}{0.45\linewidth}
\centering
  \includegraphics[width=0.95\textwidth]{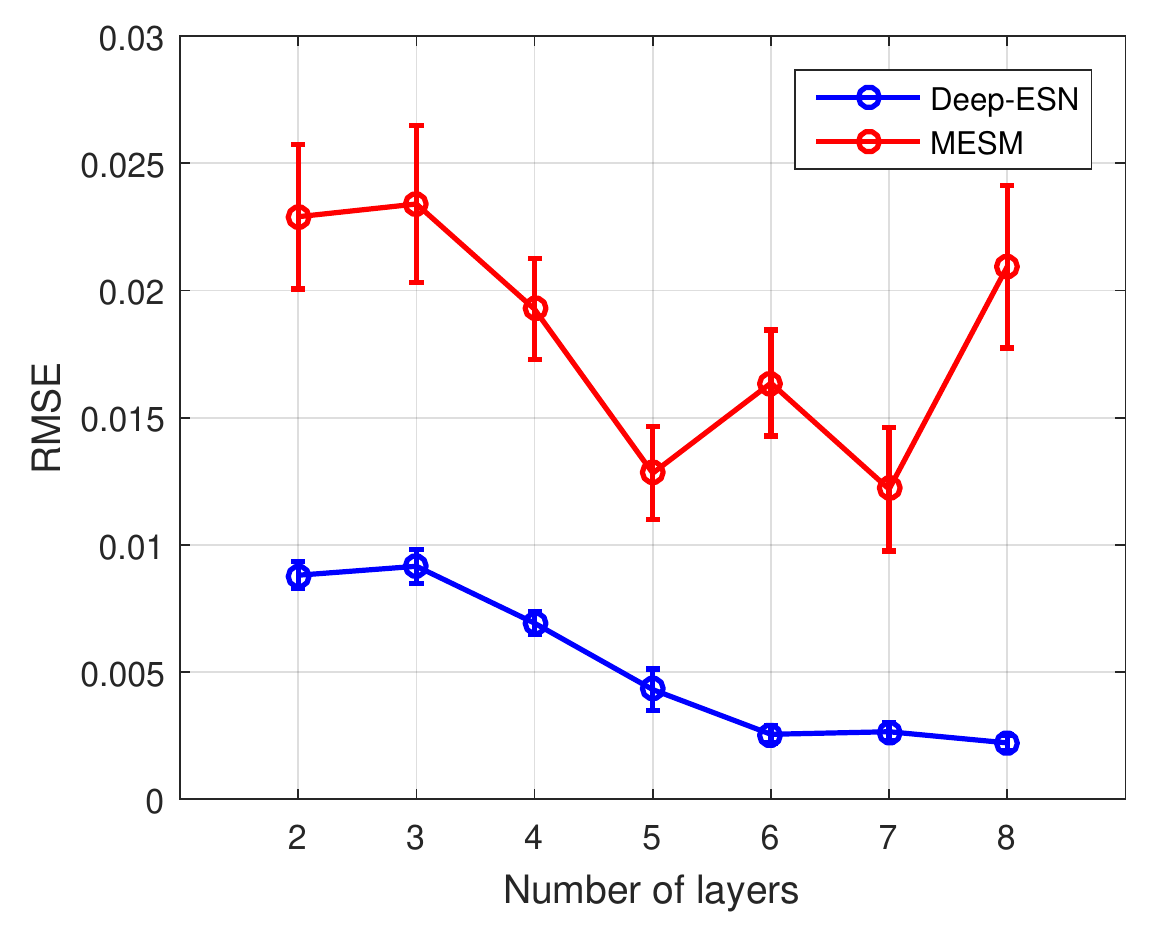}
  \centerline{(a)}
\end{minipage}
\begin{minipage}{0.45\linewidth}
\centering
  \includegraphics[width=0.95\textwidth]{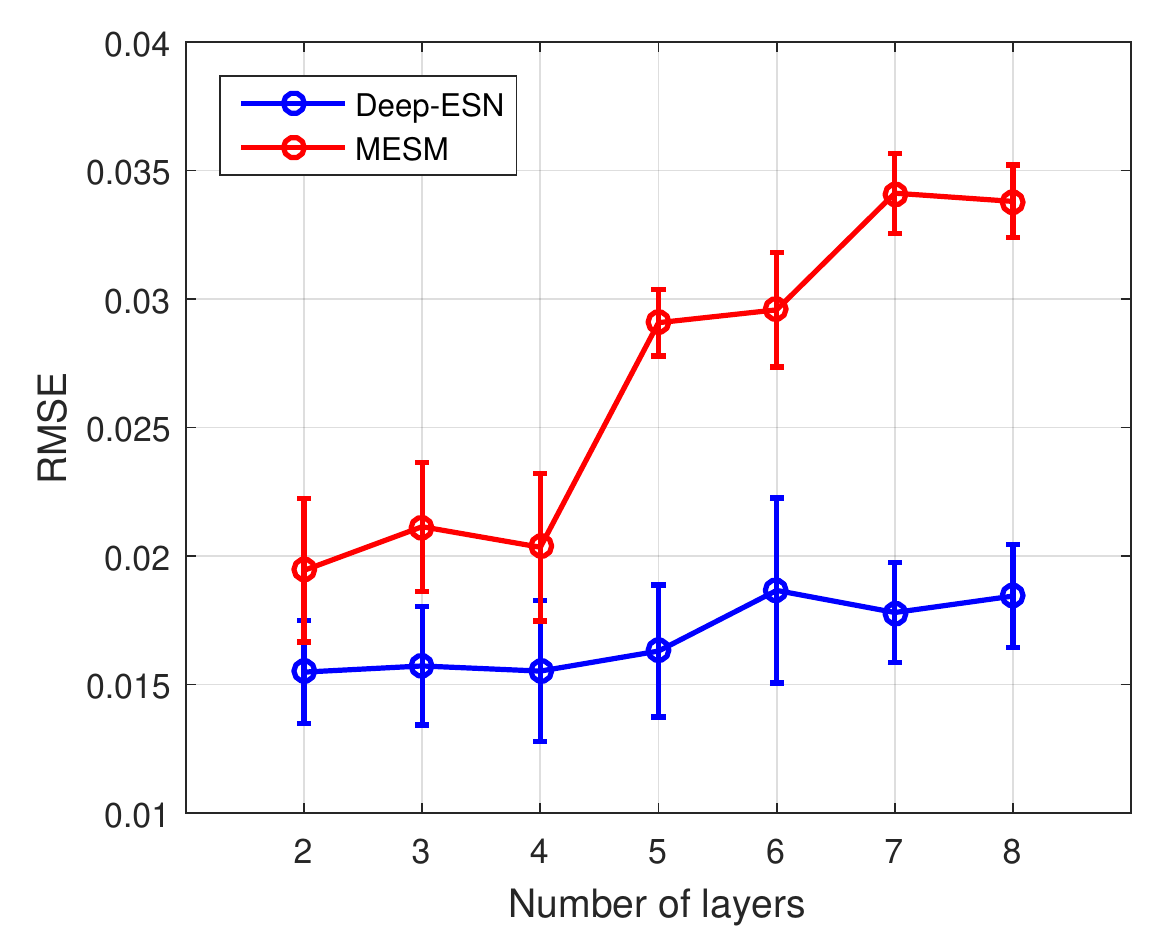}
  \centerline{(b)}
\end{minipage}
%\hfill
%\end{tabular}
\caption{Comparisons between Deep-ESN and MESM with various depths. (a) The RMSE results of the 84-step-ahead prediction of MGS; (b) The RMSE results of the one-step-ahead prediction of NARMA.}
\label{fig:depth}
\end{figure}

Third, in order to evaluate the effect of depth in the networks, we varied the number of layers in Figure~\ref{fig:depth}. We compared a Deep-ESN with PCA and  MESM on the 84-step-ahead prediction on the Mackey-Glass system, and one-step-ahead prediction of a tenth-order NARMA. The sizes of the reservoirs in all models were fixed at 300. The size of the encoder in the Deep-ESN is 30 and 200 on Mackey-Glass and NARMA, respectively.
%The Mackey-Glass prediction task requires more memory capacity, while the ten-order NARMA task requires more nonlinearity.

As shown in Figure~\ref{fig:depth}, we see that the effects of depth differ on these two tasks. In the Mackey-Glass prediction task, networks with more layers have much better performance, while in the NARMA task, the networks with only 2$\sim$4 layers perform well. This observation is consistent with the results in our previous nonlinear real world experiments (see Table \ref{results:sunspot} and \ref{results:temperature}). On the other hand, comparisons between Deep-ESN and MESM in Figure~\ref{fig:depth} also demonstrate the effectiveness of our proposed model.

\begin{figure}[t!]
\centering
\begin{minipage}{0.44\linewidth}
\centering
  \includegraphics[width=0.95\textwidth]{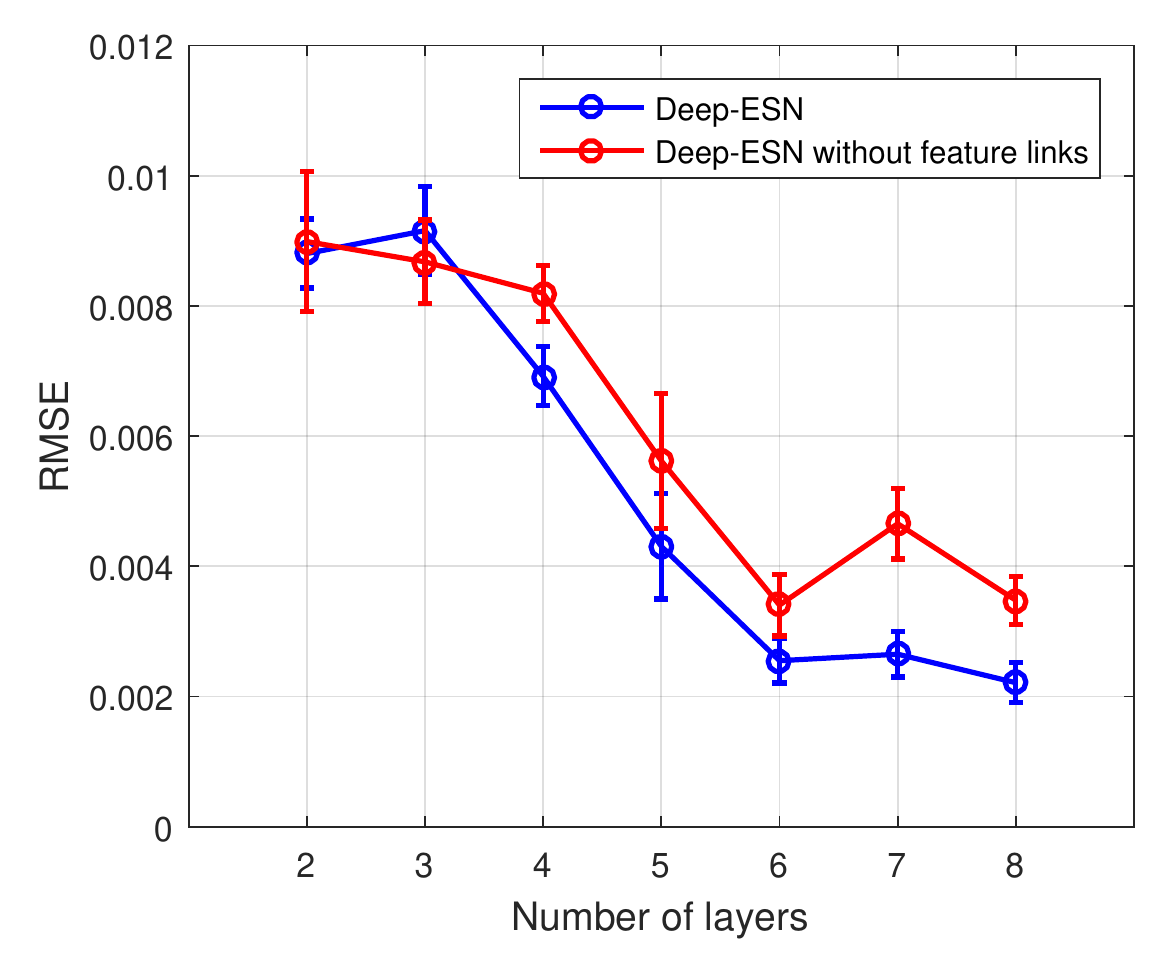}
  \centerline{(a)}
\end{minipage}
\begin{minipage}{0.44\linewidth}
\centering
  \includegraphics[width=0.95\textwidth]{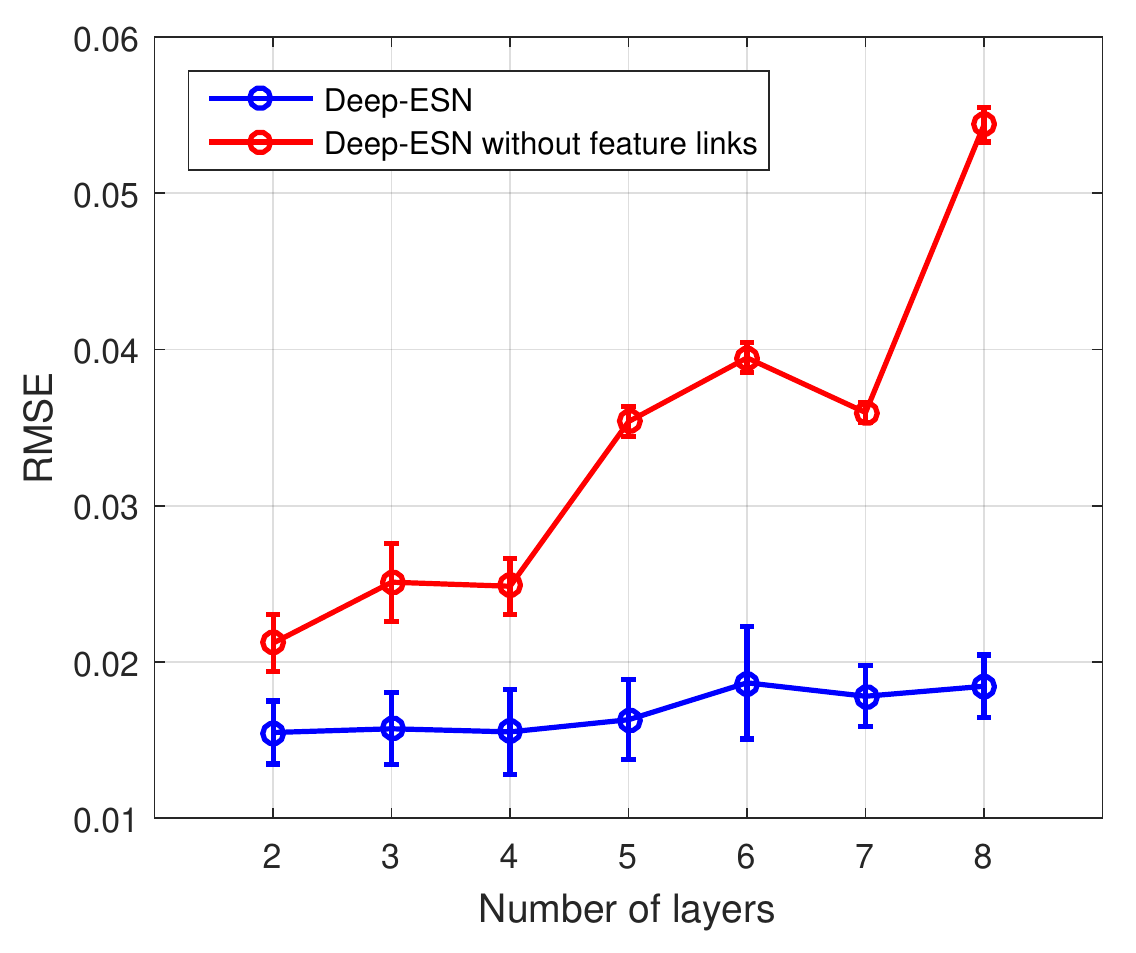}
  \centerline{(b)}
\end{minipage}
%\hfill
%\end{tabular}
\caption{Comparisons between Deep-ESNs with/without feature links. (a) The RMSE results of the 84-step-ahead prediction of MGS; (b) The RMSE results of the one-step-ahead prediction of NARMA.}
\label{fig:withoutlinks}
\end{figure}

Fourth, we investigate the effects of feature links when we vary the depth of Deep-ESNs. There are two contrast trends in Mackey-Glass (Figure~\ref{fig:withoutlinks} (a)) and NARMA (Figure~\ref{fig:withoutlinks} (b)). In the former task, networks with/without feature links both have much better performance with more layers, while in the latter task, the networks without feature links only perform well with 2$\sim$4 layers. On the other hand, when the networks are shallow, there are not much performance differences between Deep-ESN with feature links and the ones without feature links. In other words, the deeper the network is, the more useful the feature links are.

\begin{figure}[t!]
	\centering
	\includegraphics[width=0.37\textwidth]{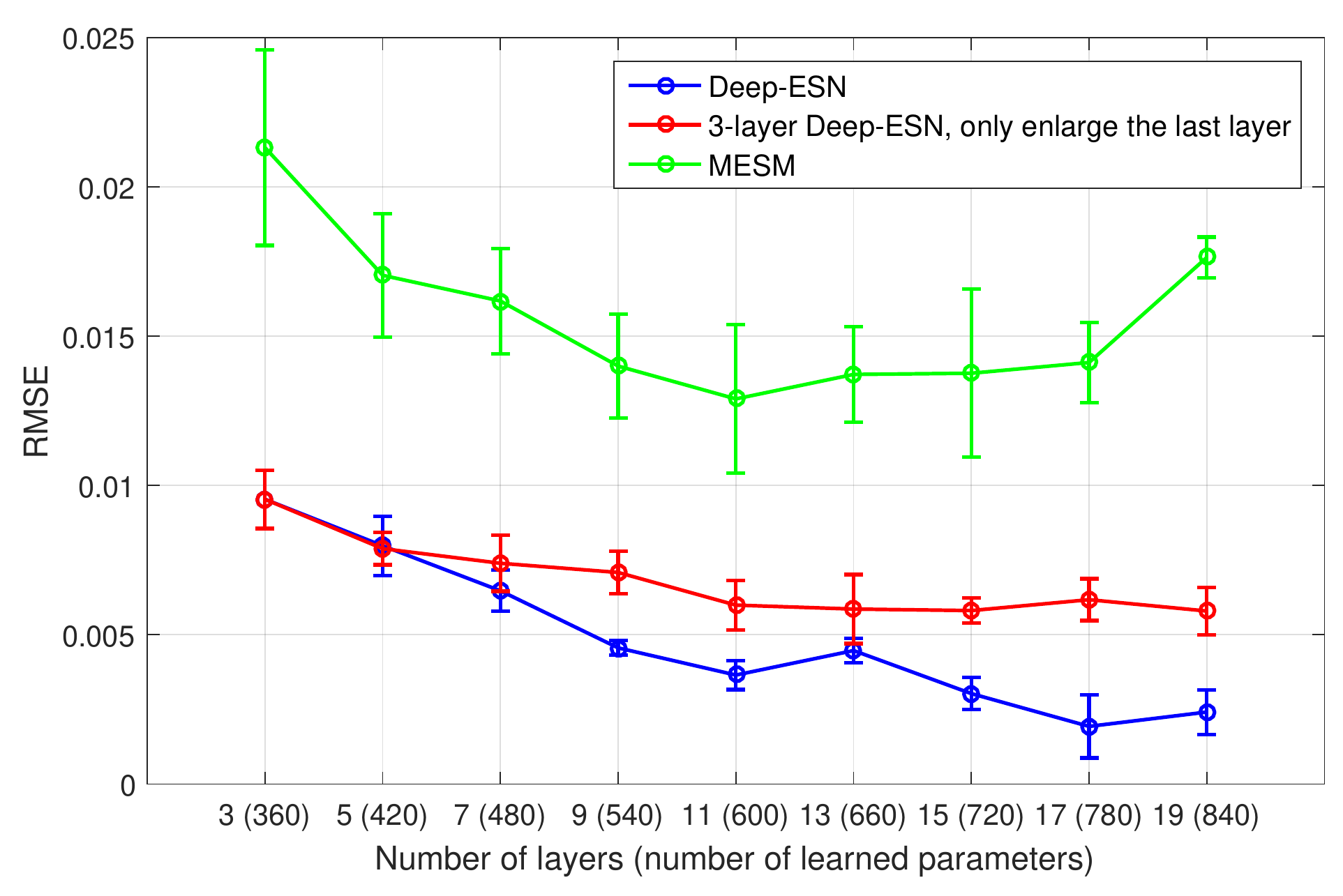}
	\caption{Adding more layers by copying the hyper-parameters in the second and third reservoirs. %{\Large GWC: There is a typo in the x-axis label: leanred -> learned}
}
	\label{auto_add_layers}
\end{figure}

Last, we explore a general schema to further deepen our Deep-ESN (adding more layers to depth 19). Since the number of  hyper-parameters (\(IS\), \(SR\) and \(\gamma\)) become unwieldy for these many layers, we first optimize a 3-layer network (Deep-ESN with PCA and MESM, each reservoir is fixed to be 300), and then directly copy the hyperparameters of the 2nd and 3rd reservoir into the $2j^{th}$ and $2(j+1)^{th}$ reservoirs, where $j\in \mathbb{Z}^+$. In this way, we can deepen our Deep-ESN without a high optimization cost. The settings of the first three layers (including other fixed parameters) are listed in Table~\ref{parms_analysis}. The performance of the networks with depth from 3 to 19 are shown in Figure~\ref{auto_add_layers}.

As seen in Figure~\ref{auto_add_layers}, it appears that this way of directly adding layers, by copying hyperparameters, is feasible in our Deep-ESN (blue line), and improves performance. The Deep-ESN obtains the best performance at 17 layers. Moreover, we also consider a shallower model Deep-ESN with only 3-layers (red line), where we do not add any more layers, but instead enlarge its last reservoir with the same number of learned output weights (numbers in the brackets) as the Deep-ESN with the corresponding depth. Compared with this model (red line), the Deep-ESN outperforms it with more and more layers added. It also verifies the Deep-ESN is better than shallow models with the same scale of learned weights. For the MESM (green line), we see that this method of deepening it has a sweet spot at 11 layers, but performance worsens after that. In general, the Deep-ESN is a better and more efficient hierarchical choice.

\subsection{Collinearity Analysis}

%\begin{figure*}[ht!]
%\centering
%\begin{minipage}{0.22\linewidth}
%\centering
%  \includegraphics[width=0.99\textwidth,height=1.4in]{figures/multiscale-MESM}
%  \centerline{(a) MESM}
%\end{minipage}
%%\hfill
%\begin{minipage}{.22\linewidth}
%\centering
%  \includegraphics[width=0.99\textwidth,height=1.4in]{figures/multiscale-PCA}
%  \centerline{(b) Deep-ESN with PCA}
%\end{minipage}
%\begin{minipage}{0.22\linewidth}
%\centering
%  \includegraphics[width=0.99\textwidth,height=1.4in]{figures/multiscale-ELMAE}
%  \centerline{(c) Deep-ESN with ELM-AE}
%\end{minipage}
%%\hfill
%\begin{minipage}{0.22\linewidth}
%\centering
%  \includegraphics[width=0.99\textwidth,height=1.4in]{figures/multiscale-RP}
%  \centerline{(d) Deep-ESN with RP}
%\end{minipage}
%%\end{tabular}
%\caption{Visualization of the multiscale dynamics in four hierarchical models: MESM, and Deep-ESNs with the three encoder variants.}
%\label{fig:multiscale_dynamics}
%\end{figure*}

The high-dimensional projection afforded by the echo-state reservoir is an important feature of reservoir computing. Through this process, a reservoir network can produce abundant echo states in a high-dimensional space, enhancing the separability of samples, and allowing the outputs to be obtained by a simple linear combination of these variables. However, solving this regression problem has a collinearity issue due to redundant echo states and variables (components) in those echo states. Redundant states lead to the inability to make different predictions when the states are similar, while redundant components waste predictor variables. It is better for the echo states to have as many independent contributions as possible.
%While it can not be guaranteed, it would be best if the variables are as uncorrelated as possible, which PCA guarantees.
Our Deep-ESN added unsupervised encoders (PCA, ELM-AE or RP) to the reservoirs. With these encoders, the model can extract uncorrelated subspace representations of echo states so as to reduce the redundancy of each reservoir. In this section, we visualize the effects of these encoders on the collinearity problem.

%There are many tools to solve this ill-posed problem, e.g., \textit{Thikhonov} regularization (\ref{eq:train_weights}).
%However, this problem has not been considered in the existing hierarchical reservoir-computing frameworks. Such as MESM \cite{MESM}, it only using the same size of reservoirs in its hierarchical direction, and do not consider this problem in its intermediate reservoirs.

\begin{figure}[t!]
	\centering
	\includegraphics[width=0.4\textwidth]{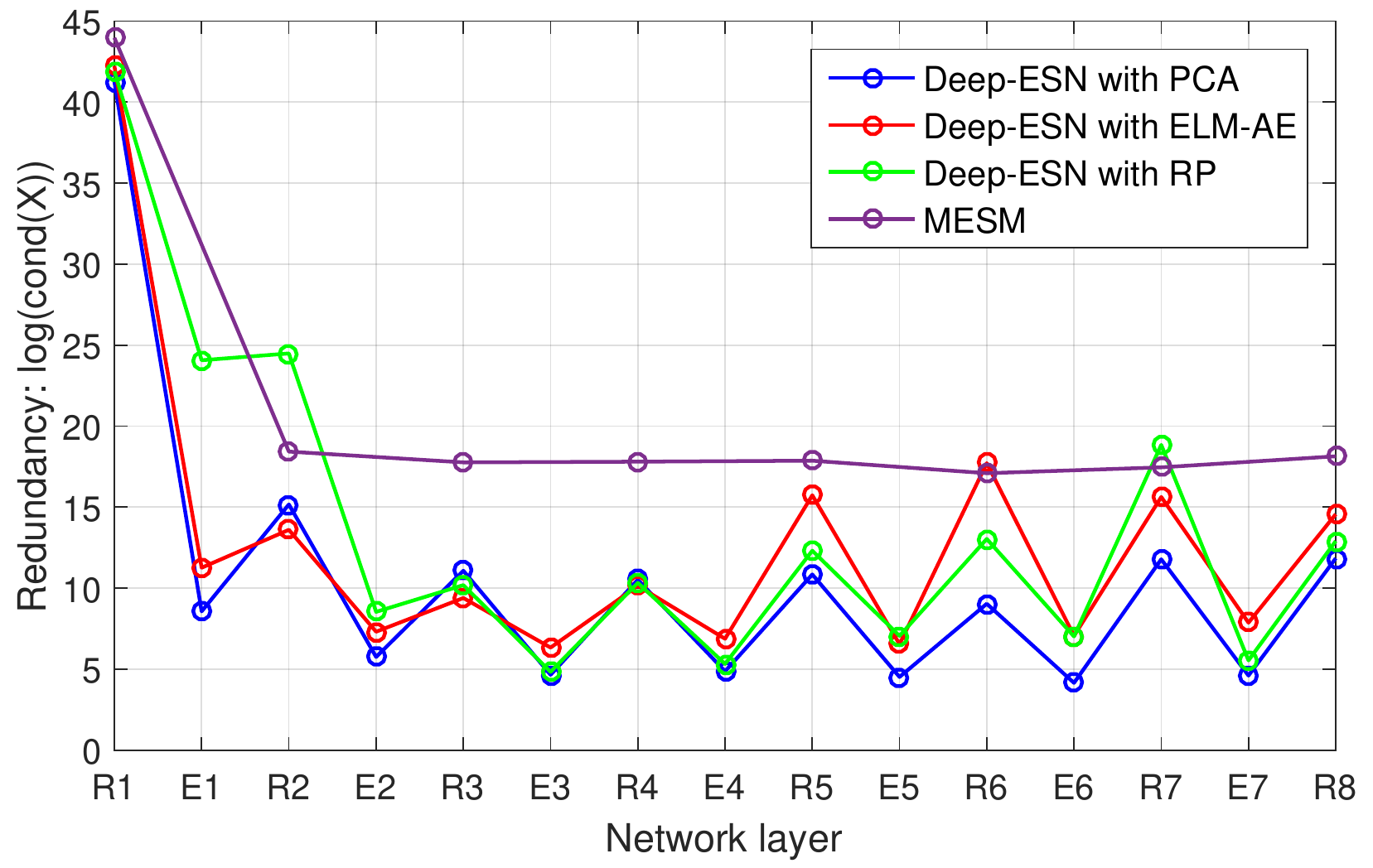}
	\caption{The results of collinearity analysis on three 8-layer Deep-ESNs and a 8-layer MESM for the MGS time series, where $R_i$ denotes the $i$-th reservoir and $E_j$ denotes the $j$-th encoder ($i=1,\dots,8, j=1,\dots,7$). In order to facilitate visualization, the logarithmic form of condition numbers are given in this plot (Y-axis).}
	\label{collinearity}
\end{figure}

Here, we use the condition number to describe the redundancy in a (sub)system. The condition number is the standard measure of ill-conditioning in a matrix \cite{Fildes1993}. Given a regression problem $\textbf{A}\textbf{x}=\textbf{b}$, where $\textbf{x}\in \mathbb{R}^n$, $\textbf{b}\in \mathbb{R}^n$, and $\textbf{A}\in \mathbb{R}^{m\times n}$, we have condition number of matrix $\textbf{A}$ defined by
\begin{equation}{}
cond(\textbf{A})=\frac{\sigma_{\max}{(\textbf{A})}}{\sigma_{\min}{(\textbf{A})}}
\end{equation}
where $\sigma_{\max}{(\textbf{A})}$ and $\sigma_{\min}{(\textbf{A})}$ are maximal and minimal singular values of $\textbf{A}$ respectively.
 %{\Large GWC: But what if the matrix (as it is in this case) is not square? Do you mean pseudoinverse? Another way to say it is that we will use the ratio of the highest and lowest singular values of the SVD of the matrix, which is another way to define the condition number. [The rest of this comment is somewhat redundant with the one above...] I think that you should say a little more here about why we care about the collinearity problem. Collinearity doesn't prevent you from doing linear regression, but what we would like is for the echo states to have as many independent contributions as possible. While we can't guarantee that, it would be best if the variables are as uncorrelated as possible, which PCA guarantees. Another point you could make here is that a high condition number lowers the accuracy of the linear regression.}
For our Deep-ESN, we compute the condition number of the echo-state collected matrix $\textbf{X}_{res}^{(i)}\in \mathbb{R}^{(N^{(i)}\times T)}$ in the $i$-th reservoir, and of the encoded-state collected matrix $\textbf{X}_{enc}^{(j)}\in \mathbb{R}^{(M^{(j)}\times T)}$ in the $j$-th encoder. For the MESM, we only compute its condition number of $\textbf{X}_{res}^{(i)}$. The results of collinearity analysis on three 8-layer Deep-ESNs with different encoders and an 8-layer MESM are shown in Fig.\ref{collinearity}.

As seen in Fig.\ref{collinearity}, for all the methods, the largest redundancy occurs in the first reservoir. With the multiple state transitions in its hierarchical direction, MESM does not reduce redundancy any further after two layers. A high condition number in MESM lowers the accuracy of the linear regression. Compared with MESM, our Deep-ESN works well with its encoders. We can see that the higher the layer, the redundancy on reservoirs will be less, especially with PCA, although this flattens out after E2 or E3. In fact, after R4, the condition number appears to increase for ELM and RP, which suggests why the PCA encoder is more effective.

Note that the condition number oscillates between encoder and reservoir layers, because PCA automatically reduces the condition number due to the orthogonalization of the data. On the other hand, we also see that the high-dimensional projection results in abundant but redundant features, which demonstrates the tradeoff between the high-dimensional projection and the hierarchical construction of reservoir computing. The high-dimensional projection is a major feature of reservoir computing. If we hope to retain this projection capacity in a hierarchical framework of reservoir computing, the size of the higher reservoir will have to increase, and its redundancy will be more serious. The PCA layer holds this in check, preventing the network from becoming overly redundant. In this way, we believe that the method of alternating projection and encoding scheme in our Deep-ESN is an effective strategy to construct a hierarchical reservoir computing framework.

\begin{figure*}[t]
	\begin{tabular}{ccccc}
		\renewcommand\arraystretch{0.8}
		&\small MESM&\small Deep-ESN with PCA&\small Deep-ESN with ELM-AE&\small Deep-ESN with RP\\
		\rotatebox{90}{\small \ }&
		\begin{minipage}{0.22\linewidth}
			\centering
			\includegraphics[width=0.99\textwidth,height=1.2in]{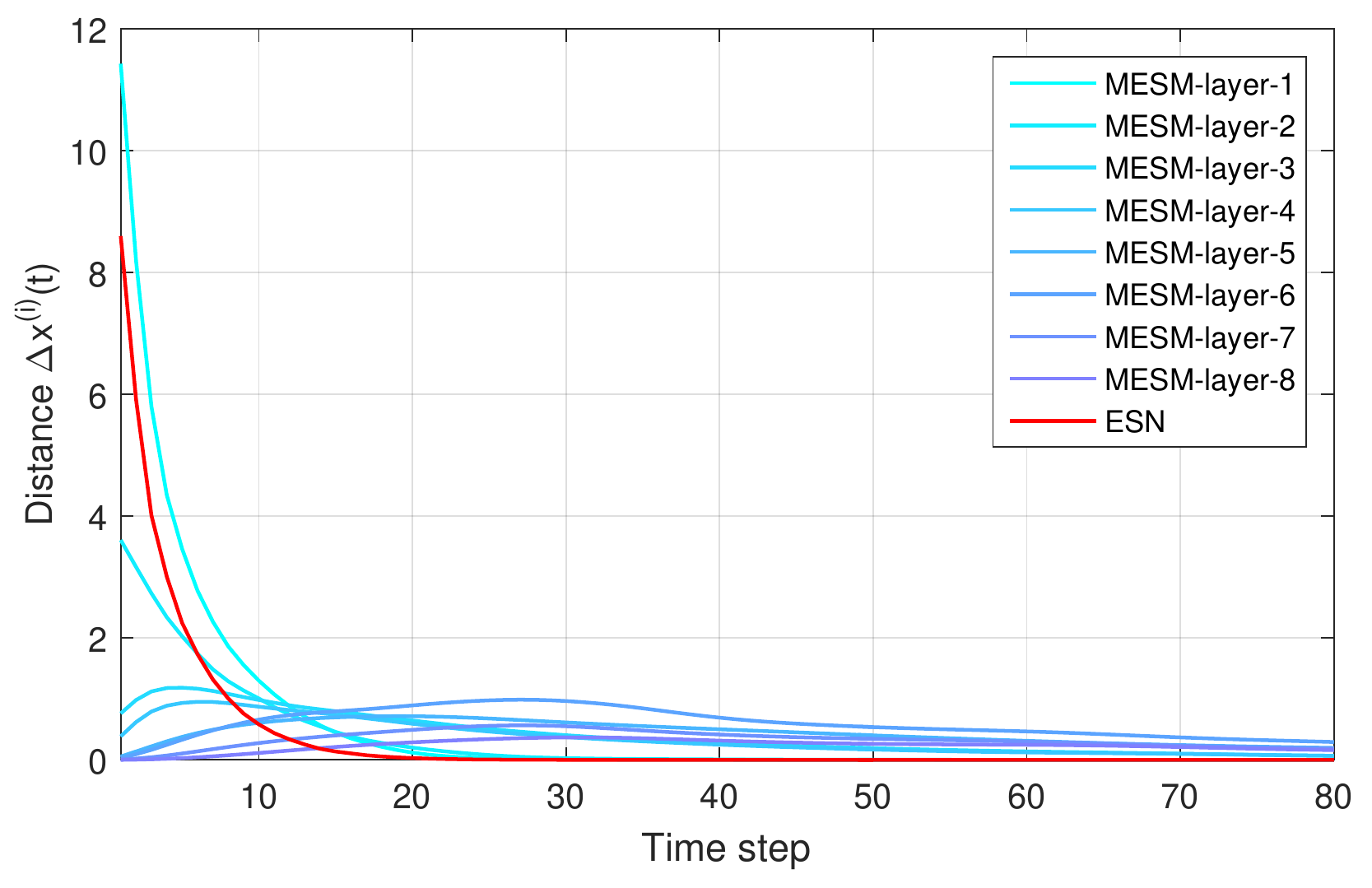}
			\centerline{\small (a)}
		\end{minipage}&
		\begin{minipage}{0.22\linewidth}
			\centering
			\includegraphics[width=0.99\textwidth,height=1.2in]{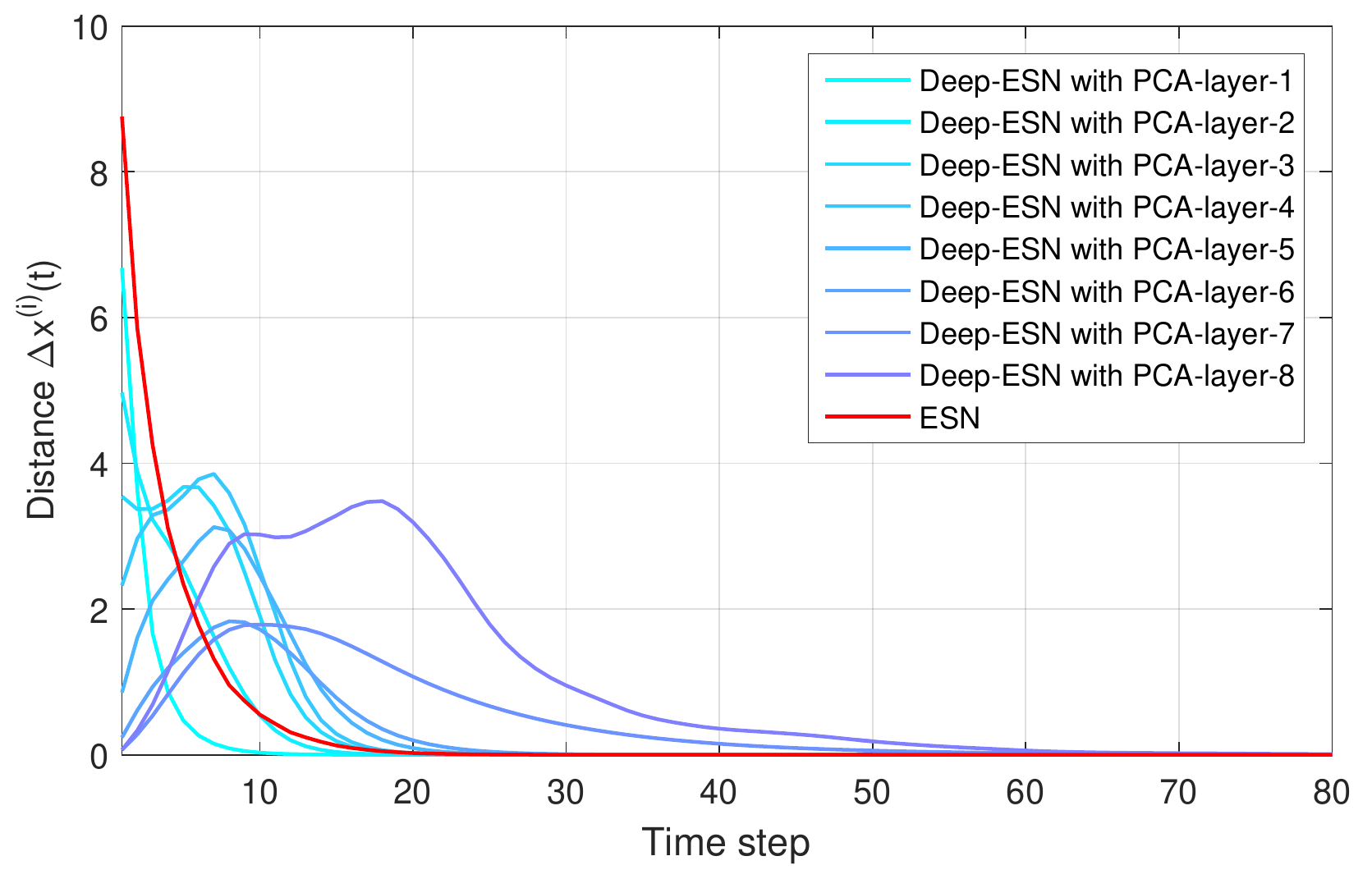}
			\centerline{\small (b)}
		\end{minipage}&
		\begin{minipage}{0.22\linewidth}
			\centering
			\includegraphics[width=0.99\textwidth,height=1.2in]{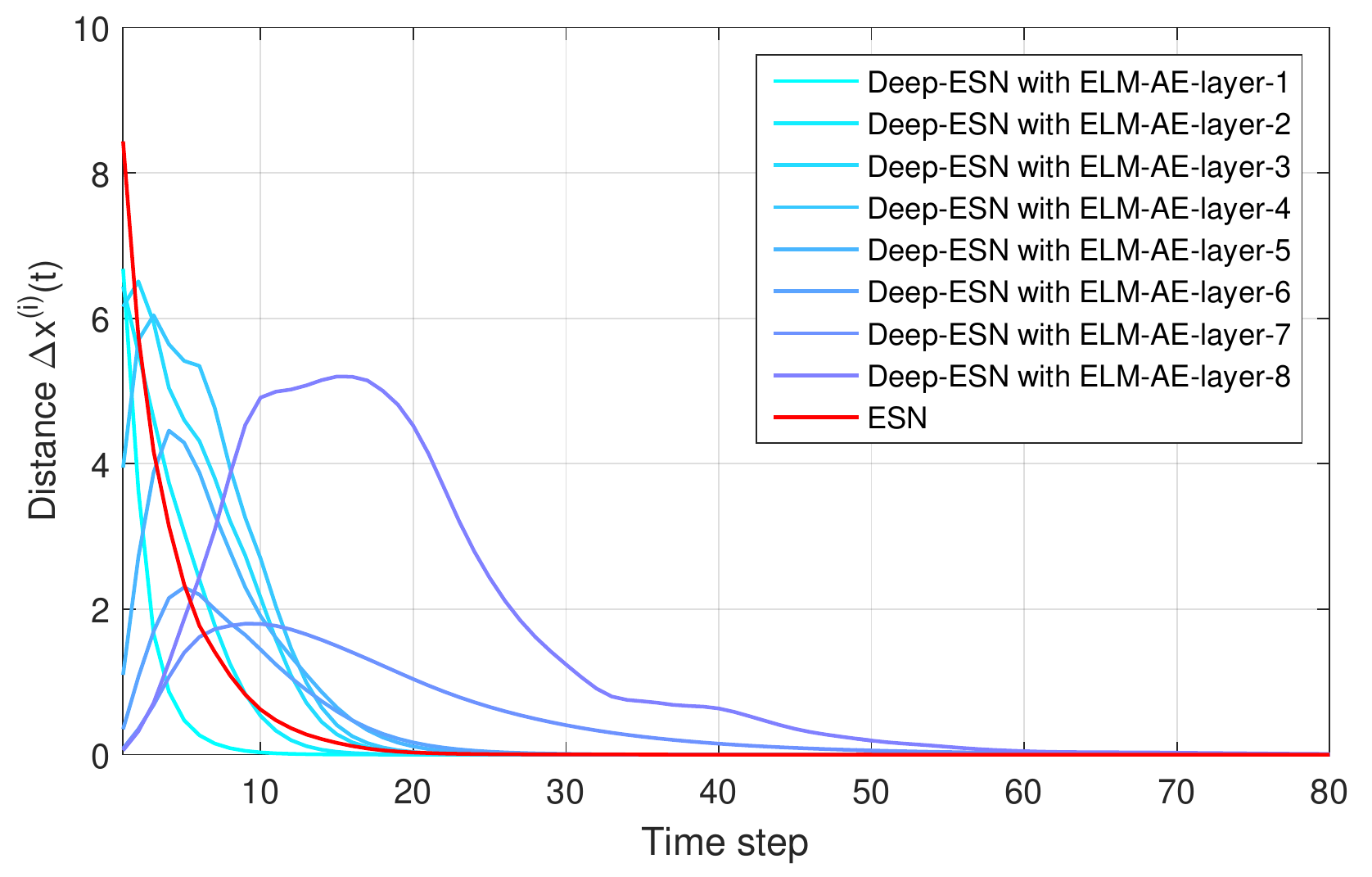}
			\centerline{\small (c)}
		\end{minipage}&
		\begin{minipage}{0.22\linewidth}
			\centering
			\includegraphics[width=0.99\textwidth,height=1.2in]{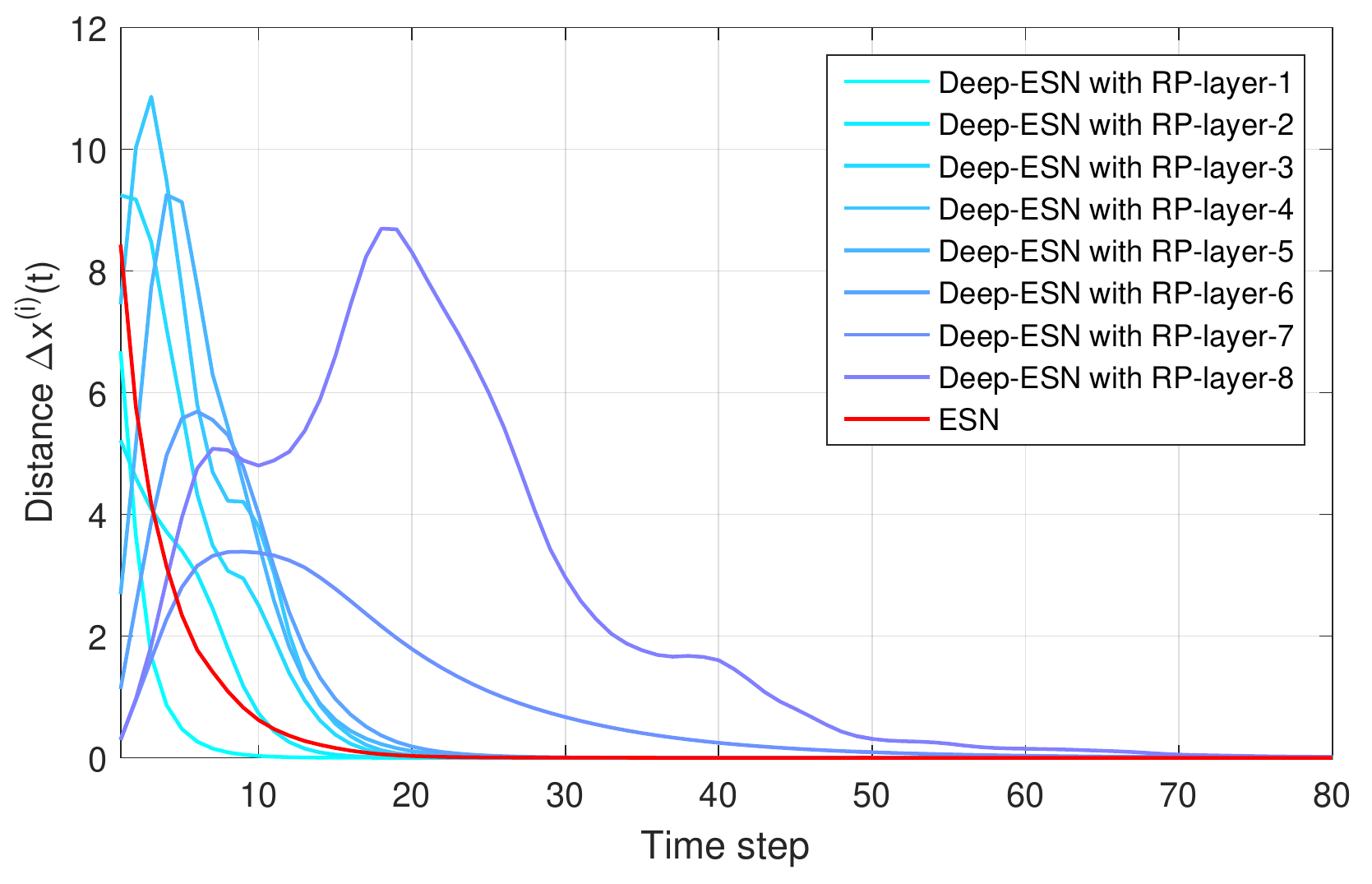}
			\centerline{\small (d)}
		\end{minipage}\\
	\end{tabular}
	\caption{Visualization of the multiscale dynamics in four hierarchical models: MESM, and Deep-ESNs with PCA, ELM-AE, and RP on Mackey-Glass.}
	\label{fig:multiscale_dynamics}
\end{figure*}

\begin{figure*}[t]
\begin{tabular}{ccccc}
\renewcommand\arraystretch{0.8}
    &\small Mackey-Glass&\small NARMA&\small Sunspots&\small Temperature\\
    \rotatebox{90}{\small Deep-ESN}&
    \begin{minipage}{0.22\linewidth}
    \centering
      \includegraphics[width=0.99\textwidth]{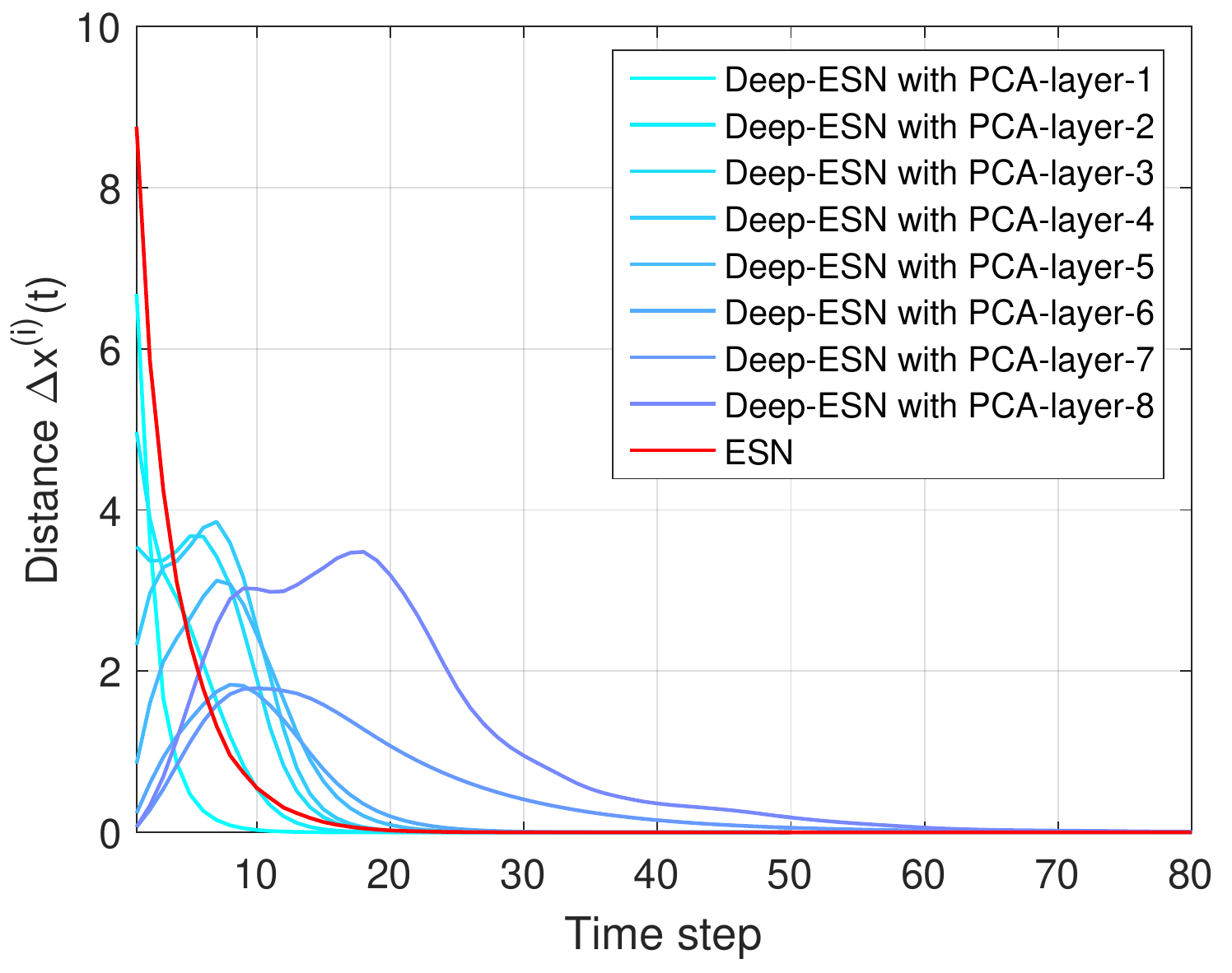}
      \centerline{\small (a1)}
    \end{minipage}&
    \begin{minipage}{0.22\linewidth}
    \centering
      \includegraphics[width=0.99\textwidth]{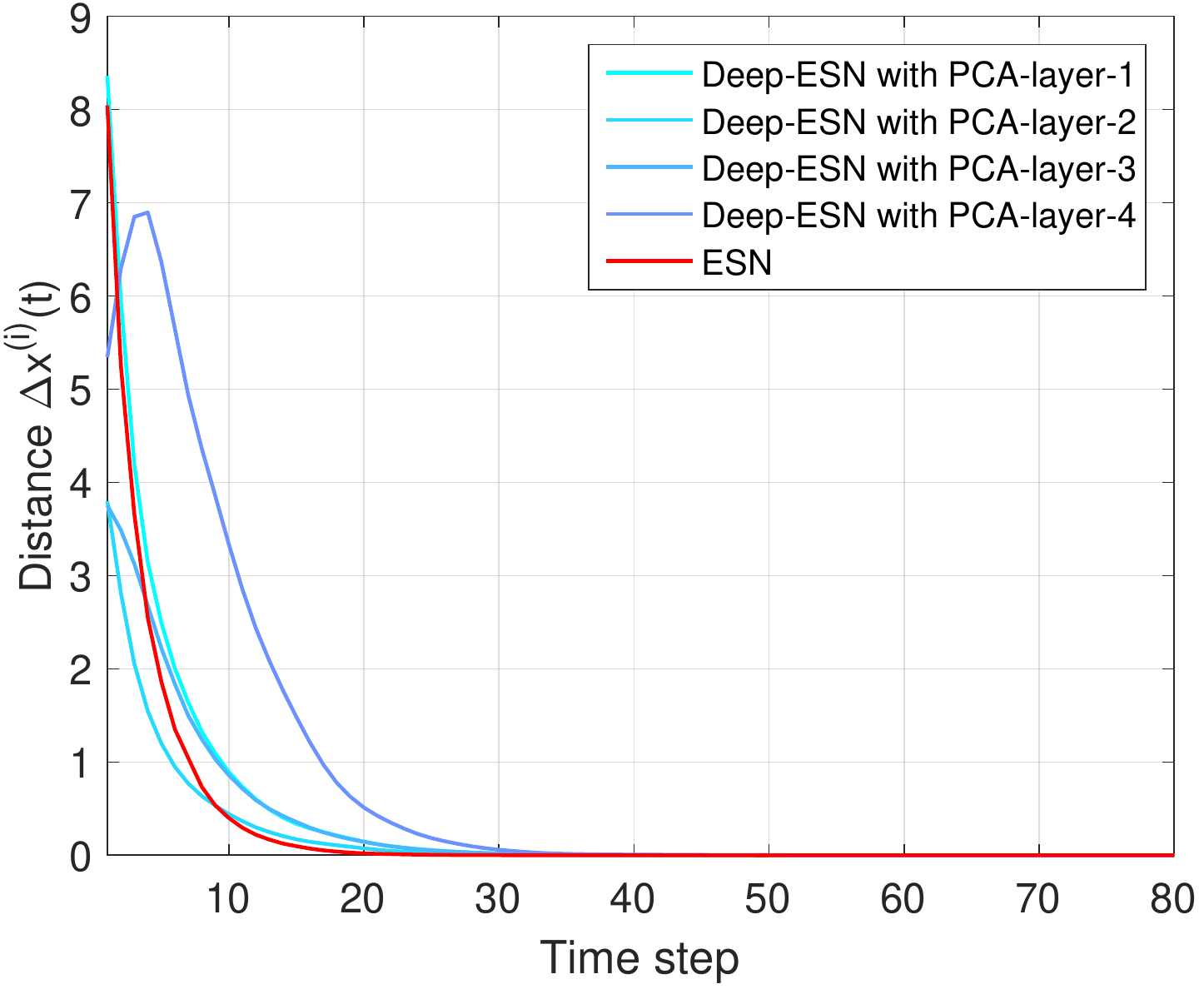}
      \centerline{\small (a2)}
    \end{minipage}&
    \begin{minipage}{0.22\linewidth}
    \centering
      \includegraphics[width=0.99\textwidth]{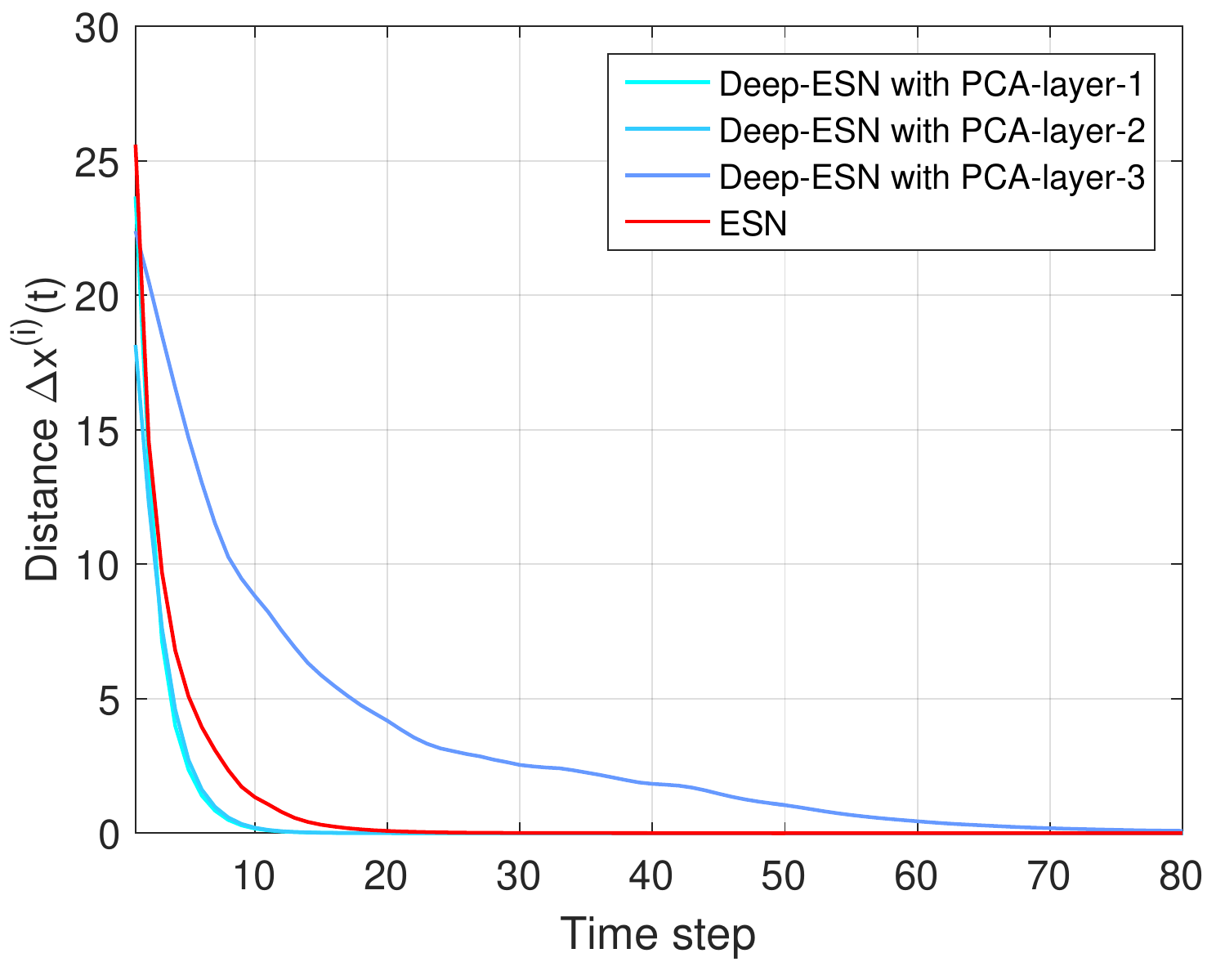}
      \centerline{\small (a3)}
    \end{minipage}&
    \begin{minipage}{0.22\linewidth}
    \centering
      \includegraphics[width=0.99\textwidth]{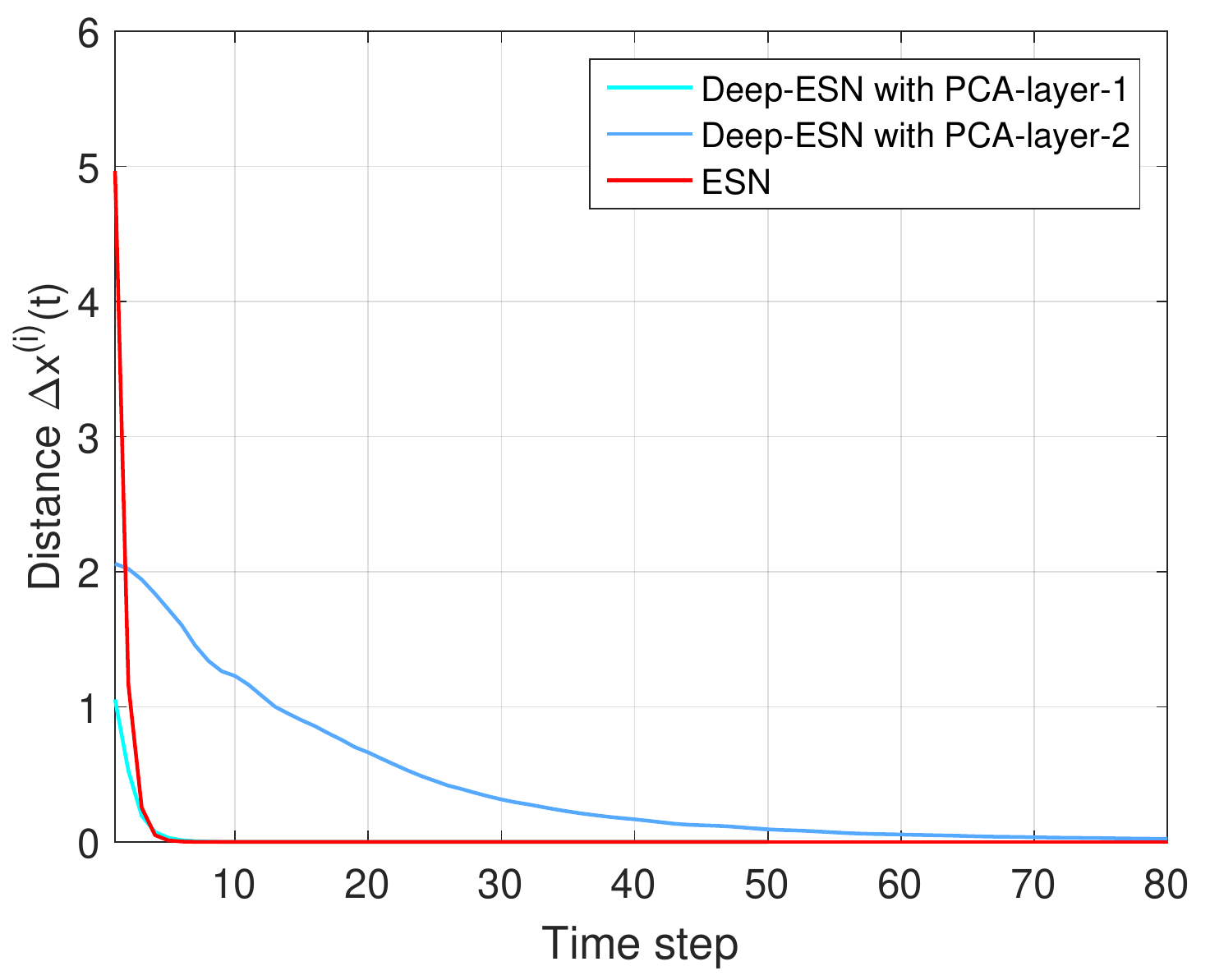}
      \centerline{\small (a4)}
    \end{minipage}\\

    \rotatebox{90}{\small MESM}&
    \begin{minipage}{0.22\linewidth}
    \centering
      \includegraphics[width=0.99\textwidth]{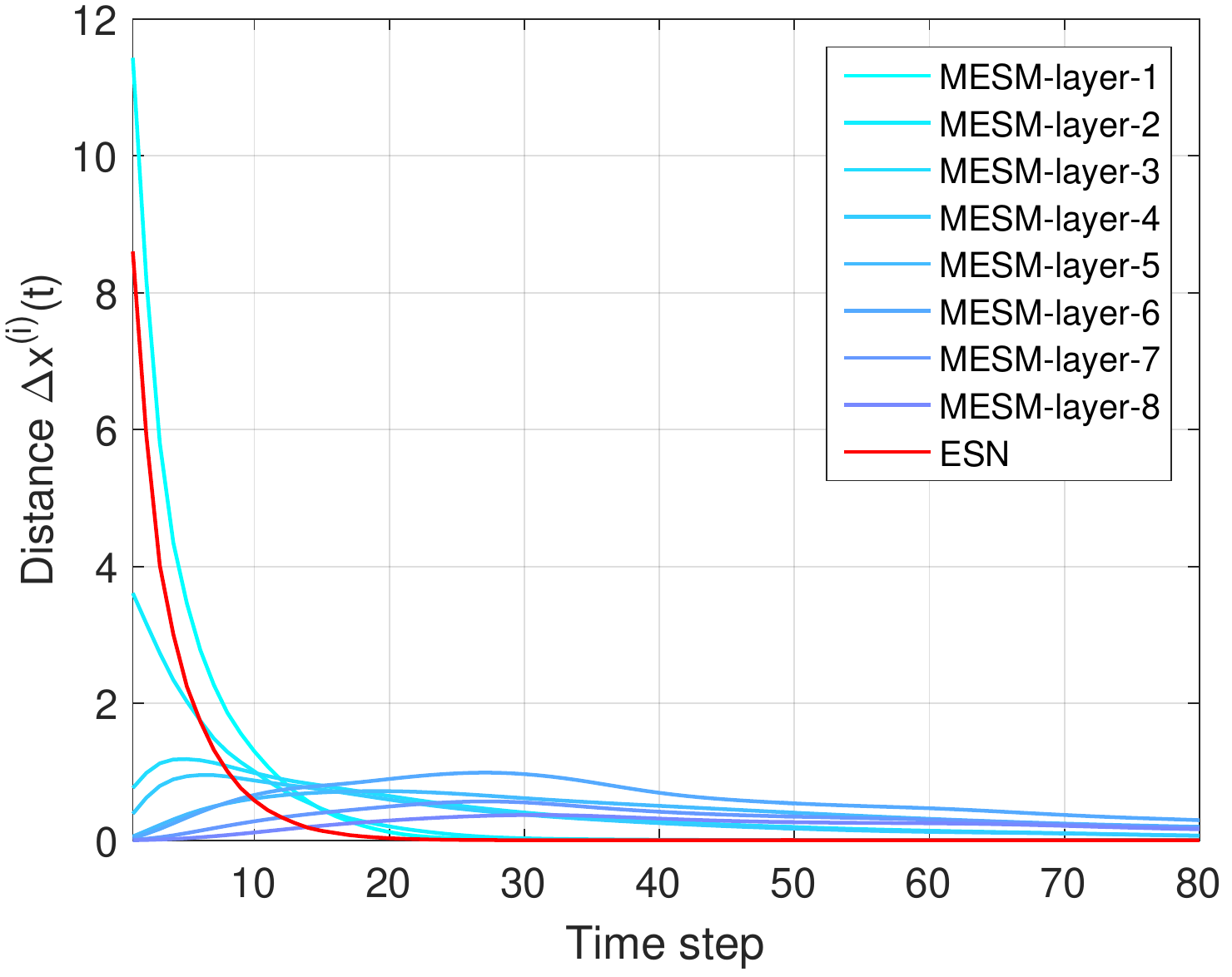}
      \centerline{\small (b1)}
    \end{minipage}&
    \begin{minipage}{0.22\linewidth}
    \centering
      \includegraphics[width=0.99\textwidth]{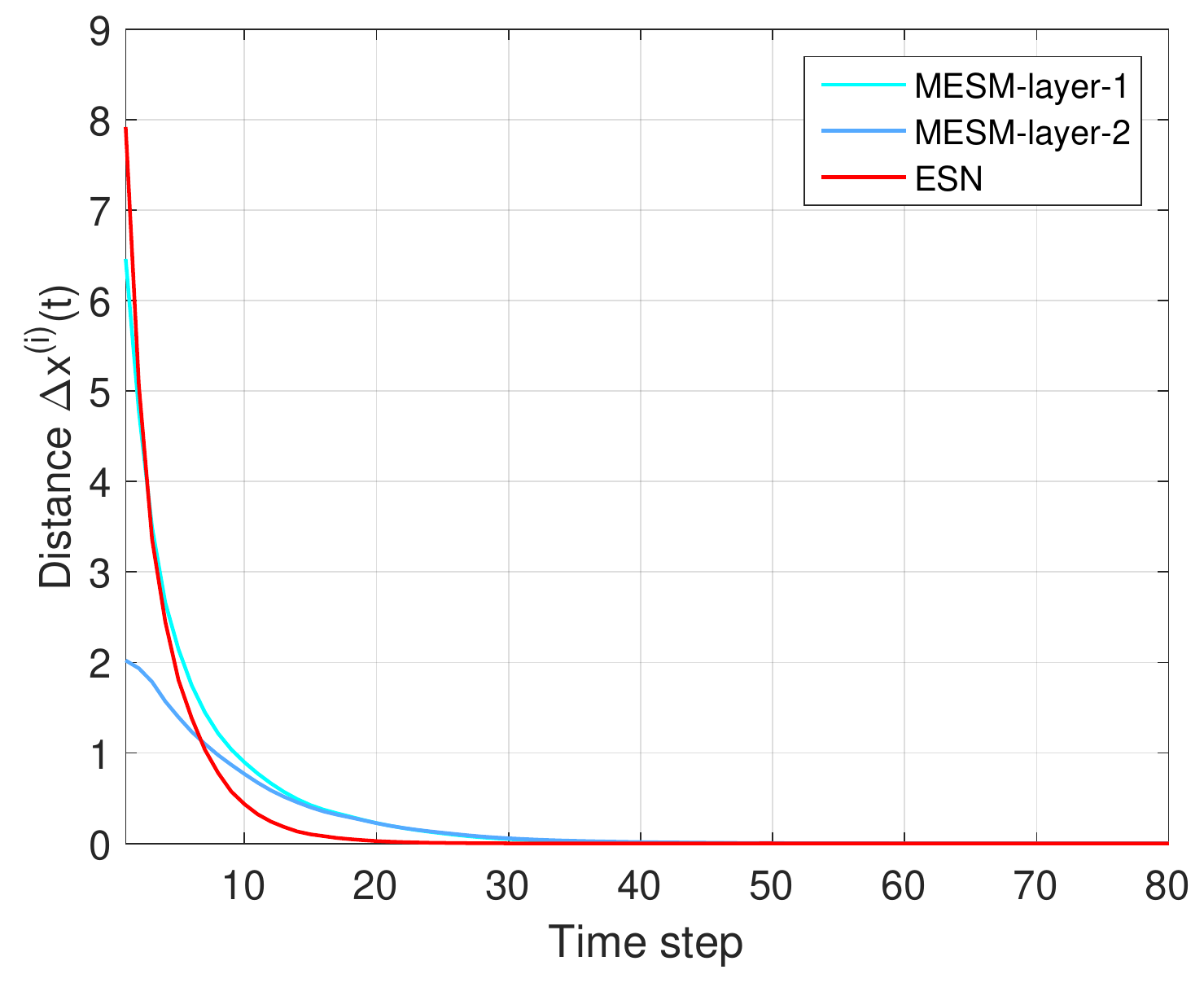}
      \centerline{\small (b2)}
    \end{minipage}&
    \begin{minipage}{0.22\linewidth}
    \centering
      \includegraphics[width=0.99\textwidth]{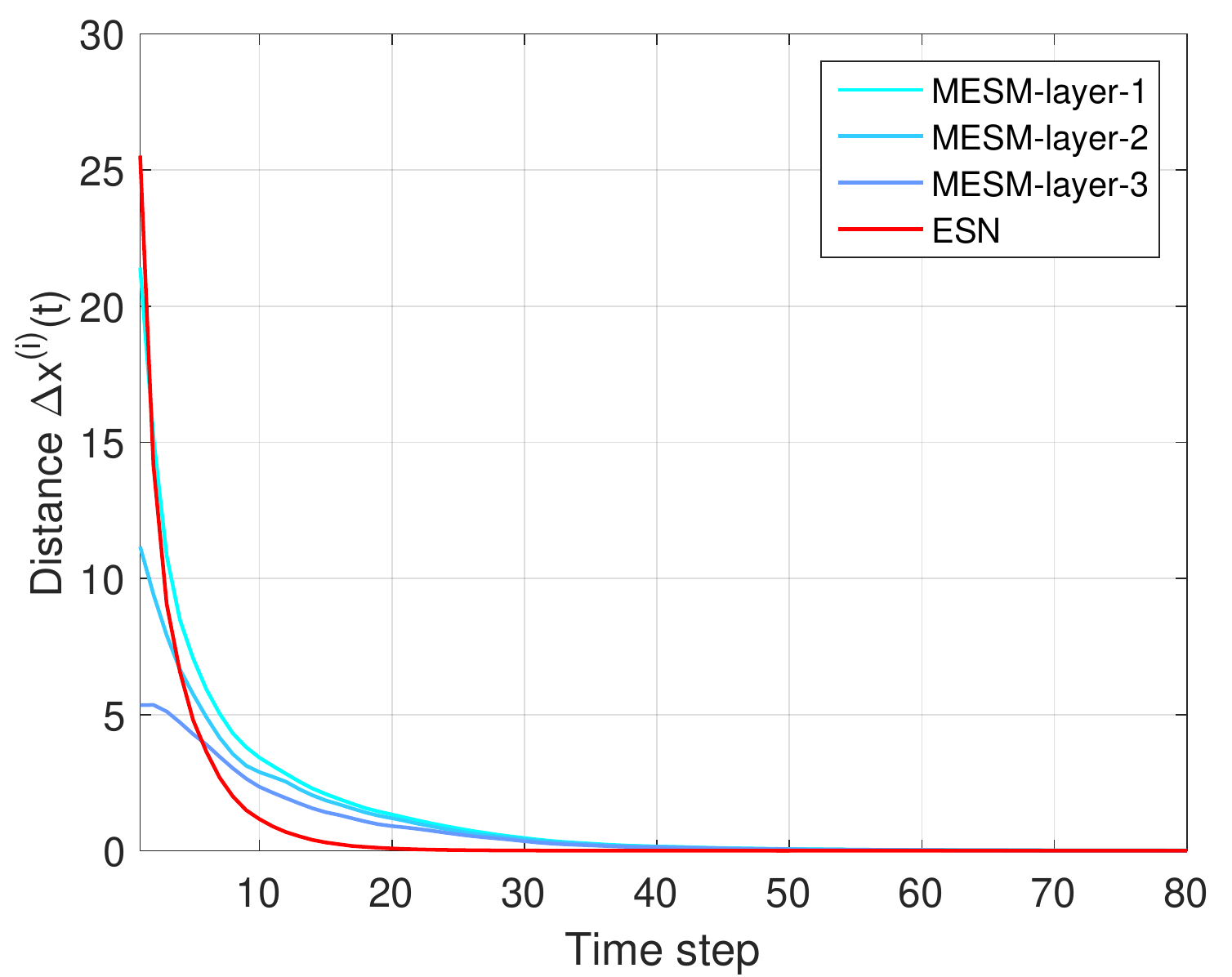}
      \centerline{\small (b3)}
    \end{minipage}&
    \begin{minipage}{0.22\linewidth}
    \centering
      \includegraphics[width=0.99\textwidth]{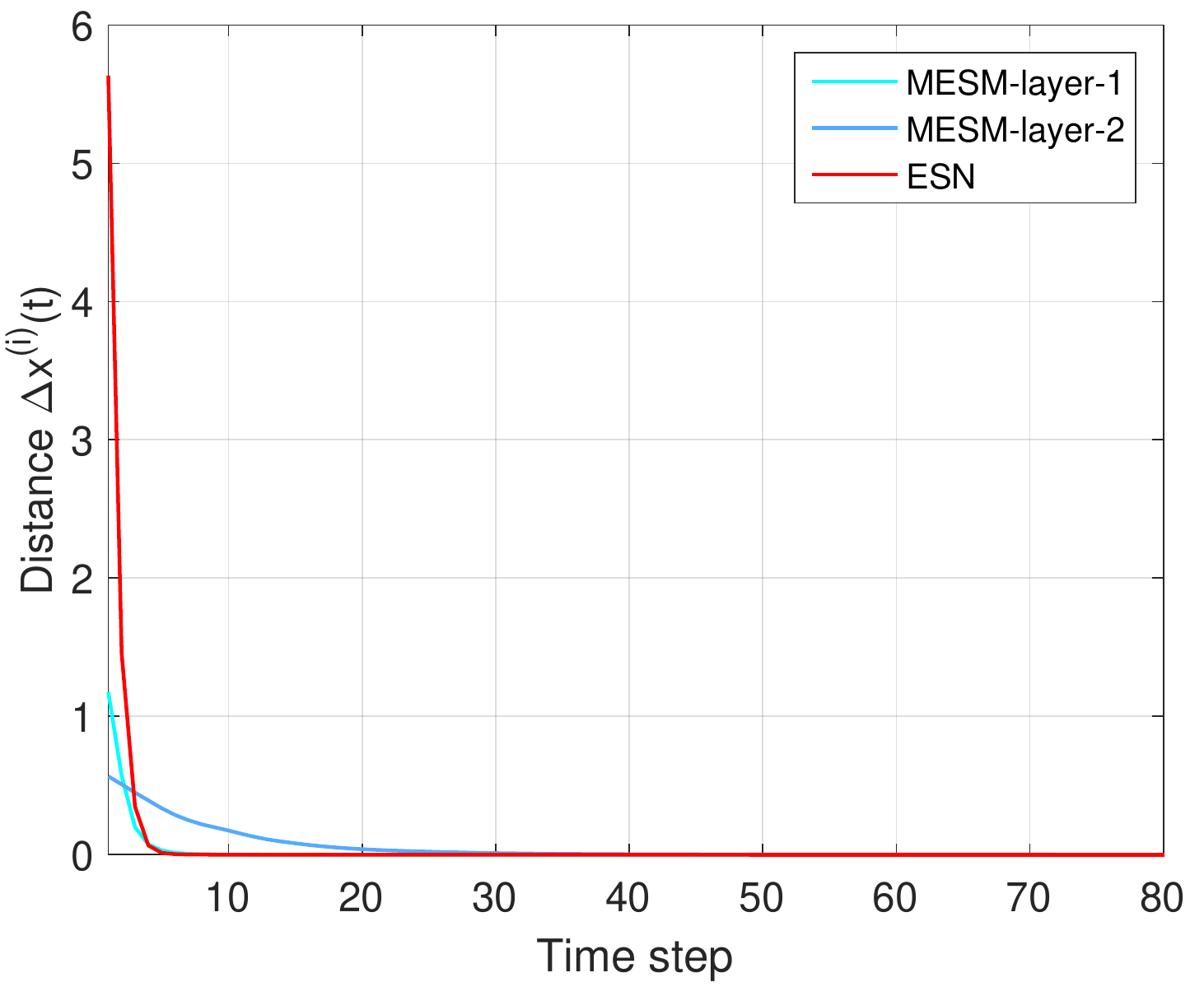}
      \centerline{\small (b4)}
    \end{minipage}
\end{tabular}
\caption{Visualization of the multiscale dynamics in MESM and Deep-ESNs with PCA on four datasets: the Mackey-Glass, NARMA, the monthly sunspot series and the daily minimum-temperature series.}
\label{fig:multiscale_dynamics_datasets}
\end{figure*}

\subsection{Visualization of Multiscale Dynamics}
\label{subsec:multiscale}

Visualizing multiscale dynamics is one of most intuitive and effective ways to understand the internal mechanism of a given hierarchical system \cite{NIPS2013_5166}. In this section, we construct an experiment similar to the work of \cite{NIPS2013_5166} to visualize the multiscale dynamics of our proposed Deep-ESN with various encoders and that of the MESM. We generate a Mackey-Glass time series $S$ with the length of 300. We then  perturb $S$ by adding a noise value
%of 1
to the time series at time step $200$, which is denoted as $S'$.
%{\Large Do you literally mean you added 1.0 to the time series? This is very high, given that the series runs between roughly -0.6 and 0.4.}
We then drive the same hierarchical ESN-based model with $S$ and $S'$, and then measure the Euclidean distance between the echo states generated by $S'$ and $S$. Formally, we measure the difference between $\mathbf{x}_{S'}^{(i)}(t)$ and the original echo states $\mathbf{x}_{S}^{(i)}(t)$ in $i$-th reservoir layer at the time step $t$, which we denote as $\bigtriangleup\mathbf{x}^{(i)}(t)$. In this way, we see how long the perturbation affects each layer, as a measure of the ``memory span'' of the network.

We consider an 8-layer Deep-ESN and an 8-layer MESM with the hyperparameters that were used in Section \ref{sec:mgs}. As a reference model, we also plot the dynamics of a single-layer ESN.
The visualization results of the multiscale dynamics are plotted in Figure~\ref{fig:multiscale_dynamics} (here we only plot 80 of the last 100 time steps for clarity). In these plots, the red line denotes the single-layer reference ESN, and blue lines to denote the perturbation effects at each layer. Darker colors correspond to deeper layers. In Figure~\ref{fig:multiscale_dynamics}(a), the multiscale dynamics of a 8-layer MESM are observed. We can see from this plot that the time-scales differences among layers are quite small and are almost dominated by the first reservoir, although there is a small, persistent memory. Therefore, directly stacking reservoirs of equal size has dynamics very similar to a shallow ESN (red line). This is unlikely to produce significant multiscale behaviors in the hierarchy.
%As seen in Fig.\ref{fig:multiscale_dynamics}, we can find that the perturbation effects of ESN  can fade away quickly as time passed due to the short-term memory capacity of its recurrent reservoir \cite{Jaeger2001The}. In this way, the multiscale dynamics can be represented by the scale discrepancy between the fade-away
%trajectories from differen layers.
%The core of the hierarchical work MESM is its additional state transition in the hierarchical direction, and MESM not consider specialized encoding in its hierarchical processes.

Figure~\ref{fig:multiscale_dynamics}(b) shows the multiscale dynamics of a Deep-ESN with PCA. Compared with the MESM in Figure\ref{fig:multiscale_dynamics}(a), the Deep-ESN produces greater and more persistent multiscale dynamics. The Deep-ESN has short time-scales in the more shallow layers, and the deeper layers show longer memory. In particular, the last reservoir has the longest time-scale.
%As time passed, the perturbation effects in each layer of Deep-ESN all can fade away to zero.
These results verify that the proposed hierarchical framework generates rich multiscale dynamic behaviors and is very suitable for modeling time series. Similar to the Deep-ESN with PCA in Figure~\ref{fig:multiscale_dynamics}(b), the ELM-AE based Deep-ESN also performs multiscale dynamics in Figure\ref{fig:multiscale_dynamics}(c). Interestingly, in the case of random projection (RP) in Figure~\ref{fig:multiscale_dynamics}(d), the distances in some subsequent layers are even larger than that in the first layer. The random projections may be enlarging the added noise in the corresponding encoder layers.
%there are some layers enlarging the perturbation discrepancy (the discrepancies from these layers are more larger than the red-line reference ESN).

Furthermore, we explore the difference of multiscale structures between four time series data: Mackey-Glass, NARMA, Sunspots, and Temperature in Figure~\ref{fig:multiscale_dynamics_datasets}. We use Deep-ESNs with optimal depth according to the results in previous sections. Similar to Figure~\ref{fig:multiscale_dynamics}, we add noise value to each time series %(1 to NARMA and Temperature, 10 to Sunspots)
and see how long the perturbation affects each layer. From Figure~\ref{fig:multiscale_dynamics_datasets}, we can see Mackey-Glass have a more multiscale structure than the other three time series. Although there are time-scales differences among layers in Deep-ESN for the other three time series, shallower Deep-ESN can capture dynamics in these datasets than in Mackey-Glass. Especially, for the Sunspots and Temperature time series, Deep-ESNs with 3 layers and 2 layers respectively can have good performances as shown in previous sections. Therefore, it is important to choose proper hierarchy of Deep-ESN for time series with various multiscale structures. Again, MESM does not show significant time-scales differences among layers no matter the time series are multiscale or not.

\section{Discussion and Conclusions}

Hierarchical multiscale structures naturally exist in many temporal data, a phenomenon that is difficult to capture by a conventional ESN. To overcome this limitation, we propose a novel hierarchical reservoir computing framework called Deep-ESNs. Instead of directly stacking reservoirs, we combine the randomly-generated reservoirs with unsupervised encoders, retaining the high-dimensional projection capacity as well as the efficient learning of reservoir computing. Through this multiple projection-encoding system, we not only alleviate the collinearity problem in ESNs, but we also capture the multiscale dynamics in each layer. The feature links in our Deep-ESN provides multiscale information fusion, which improves the ability of the network to fit the time series.

We also presented a derivation of the stability condition and the computational complexity of our Deep-ESN. The results show that our Deep-ESN with efficient unsupervised encoders (e.g., PCA) can be as efficiently learned as a shallow ESN, retaining the major computational advantages of traditional reservoir-computing networks.

In the experiments, we demonstrated empirically that our Deep-ESNs outperform other baselines, including other approaches to multiscale ESNs, on four time series (two chaotic systems and two real-world time series).
%While all three variants of our Deep-ESN consistently outperformed the other models, the PCA version performed best on nearly all benchmarks.
Furthermore, we found that increasing the size of the reservoirs generally improved performance, while increasing the size of the encoder layer showed smaller improvements. We also showed that increasing the depth of the network could either help or hurt performance, depending on the problem. This demonstrates that it is important to set the network structure parameters using cross-validation.

%We found that increasing the size of the reservoirs generally improved performance, especially for PCA encoders.
%while increasing the size of the encoder layer showed smaller improvements. We also found that increasing the size of the encoder layers and increasing the size of the last reservoir had similar effects on performance.
%Increasing the depth of the network could either help or hurt performance, depending on the problem. This demonstrates that it is important to set the network structure parameters using cross-validation. We also found that the feature links have greater effects in deeper models than shallower ones, which verifies adding feature links is an effective way to fuse multiscale dynamics obtained by each reservoir.
%Finally, by directly copying the hyperparameters of the 2nd and 3rd reservoir into the following ones, we further deepen our Deep-ESN to depth 19 and improve performance.

%% {\Large GWC: I assume you used CV to set things like depth and reservoir size - I could not find this described in the paper, but it should be!}
%

We also evaluated how the model overcomes the collinearity problem by measuring the condition numbers of the generated representations at different layers of the network. We found that using the encoders controlled this redundancy, especially in the case of PCA. On the other hand, simply stacking reservoirs as in MESM~\cite{MESM} leads to higher condition numbers overall. This suggests that the encoders are a vital part of the design of the system, and one of their main effects is to control the collinearity in deeper reservoirs.

Finally, we investigated the multiscale dynamics in our models by using a perturbation analysis. We found that all of our models demonstrated long-term memory for the perturbation, which was most evident in the final layer. The MESM seemed to never quite recover from the perturbation, with very small but persistent effects. We also found the four time series we used have different multiscale structures. Thus, the different hierarchies of Deep-ESNs can deal with various multiscale dynamics.

Reservoir computing is an efficient method to construct recurrent networks that model dynamical systems. This is in stark contrast to deep learning systems, which require extensive training. The former pursues conciseness and effectiveness, but the latter focuses on the capacity to learn abstract, complex features in the service of the task. Thus, there is a gap between the merits and weaknesses of these two approaches, and a potentially fruitful future direction is to discover a way to bridge these two models, and achieve a balance between the efficiency of one and the feature learning of the other. Our Deep-ESN is a first step towards bridging this gap between reservoir computing and deep learning.
%In the future, we will explore more efficient solutions for this direction.

% or


\appendices

%\section*{Acknowledgments}
%
%
%The work described in this paper was partially funded by the National Natural Science Foundation of China (Grant No. 61502174, 61402181), the Natural Science Foundation of Guangdong Province (Grant No. S2012010009961, 2015A030313215), the Science and Technology Planning Project of Guangdong Province (Grant No. 2016A040403046), the Guangzhou Science and Technology Planning Project (Grant No. 2014J4100006, 201704030051), the Opening Project of Guangdong Province Key Laboratory of Big Data Analysis and Processing (Grant No. 2017014), and the Fundamental Research Funds for the Central Universities (Grant No. D2153950). It was also supported by the National Science Foundation (USA) grant SMA 1041755 to the Temporal Dynamics of Learning Center, an NSF Science of Learning Center.

\ifCLASSOPTIONcaptionsoff
  \newpage
\fi



%\bibliographystyle{IEEEtran}
\bibliographystyle{IEEEtran}
\bibliography{IEEEabrv,bib/tnnls}


%\end{IEEEbiography}


%\newpage


%\vfill


%\begin{IEEEbiography}[{\includegraphics[width=1in,height=1.25in,clip,keepaspectratio]{authors/shenlifeng}}]{Lifeng Shen} is currently
%\end{IEEEbiography}

\end{document}